\documentclass[sigconf]{acmart}

\usepackage{booktabs} % For formal tables

\usepackage{amsmath}
\usepackage{url}  %Required
\usepackage{graphicx}  %Required
\usepackage{amsfonts,amssymb}
\usepackage{verbatim}
\usepackage{multirow}
\usepackage{verbatim}
\usepackage{bm}
\usepackage{color}
\usepackage{amsmath}
\usepackage{sansmath}
\usepackage{balance}
\usepackage{longtable}
\usepackage{epstopdf}

\makeatletter
\newif\if@restonecol
\makeatother

\usepackage{epsfig}
\usepackage{subfigure}
\usepackage[ruled]{algorithm2e}

\usepackage{mathrsfs}

\begin{document}

\title{Learning to Collaborate: Multi-Scenario Ranking via Multi-Agent Reinforcement Learning}

\author{Jun Feng$^{1,\text{\dag}}$, Heng Li$^{2,\text{\dag}}$,  Minlie Huang$^{1,*}$, Shichen Liu$^{2,*}$, Wenwu Ou$^{2}$, \\Zhirong Wang$^{2}$, Xiaoyan Zhu$^{1}$}

\affiliation{%
  \department{$^{1}$State Key Lab on Intelligent Technology and Systems, Tsinghua National Lab for Information Science and Technology}
  \department{$^{1}$Department of Computer Science and Technology, Tsinghua University, Beijing, China}
  \institution{$^{2}$Alibaba Group, Hangzhou, China}
}
\email{feng-j13@mails.tsinghua.edu.cn;heng.lh@alibaba-inc.com; aihuang@tsinghua.edu.cn;shichen.lsc@alibaba-inc.com;santong.oww@taobao.com; qingfeng@alibaba-inc.com; zxy-dcs@tsinghua.edu.cn}

\titlenote{Corresponding authors: Minlie Huang, aihuang@tsinghua.edu.cn; Shichen Liu: shichen.lsc@alibaba-inc.com\\
$^{\text{\dag}}$The authors contributed equally to this study.}

\copyrightyear{2018}
\acmYear{2018} 
\setcopyright{iw3c2w3}
\acmConference[WWW 2018]{The 2018 Web Conference}{April 23--27, 2018}{Lyon, France}
%\acmBooktitle{WWW 2018: The 2018 Web Conference, April 23--27, 2018, Lyon, France}
\acmPrice{}
\acmDOI{10.1145/3178876.3186165}
\acmISBN{978-1-4503-5639-8/18/04}

\begin{abstract}
Ranking is a fundamental and widely studied problem in scenarios such as search, advertising, and recommendation. However, joint optimization for multi-scenario ranking, which aims to improve the overall performance of several ranking strategies in different scenarios, is rather untouched. Separately optimizing each individual strategy has two limitations. The first one is \textbf{lack of collaboration between scenarios} meaning that each strategy maximizes its own objective but ignores the goals of other strategies, leading to a sub-optimal overall performance. The second limitation is the \textbf{inability of modeling the correlation between scenarios} meaning that independent optimization in one scenario only uses its own user data but ignores the context in other scenarios.

In this paper, we formulate multi-scenario ranking as a fully cooperative, partially observable, multi-agent sequential decision problem. We propose a novel model named Multi-Agent Recurrent Deterministic Policy Gradient (MA-RDPG) which has a communication component for passing messages, several private actors (agents) for making actions for ranking, and a centralized critic for evaluating the overall performance of the co-working actors. Each scenario is treated as an agent (actor). Agents collaborate with each other by sharing a global action-value function (the critic) and passing messages that encodes historical information across scenarios. The model is evaluated with online settings on a large E-commerce platform. Results show that the proposed model exhibits significant improvements against baselines in terms of the overall performance.

\end{abstract}




\maketitle

\section{Introduction}
Nowadays, most large-scale online platforms or mobile Apps have multiple scenarios that may involve services such as search, advertising, and recommendation. There are some well-known platforms of different kinds. Taobao is an E-commerce platform where users can search for and buy products through querying, bookmarking, or recommendation. 
Yahoo! is a comprehensive web site where users can read news, watch movies, make shopping, and more. 
%Google is a search engine where users can find various information including news, papers, images, and videos.
%
One of the common features of these services is that ranking strategy serves as a fundamental function to provide a list of ranked items to users.
Machine learning techniques have been widely applied in optimizing these ranking strategies~\cite{clark2015, yin2016ranking, li2011unbiased, DBLP:conf/kdd/LiuXOS17} to facilitate better services for search, advertising, or recommendation. 

\begin{figure}[t]
 \centering
   \includegraphics[width=0.45\textwidth]{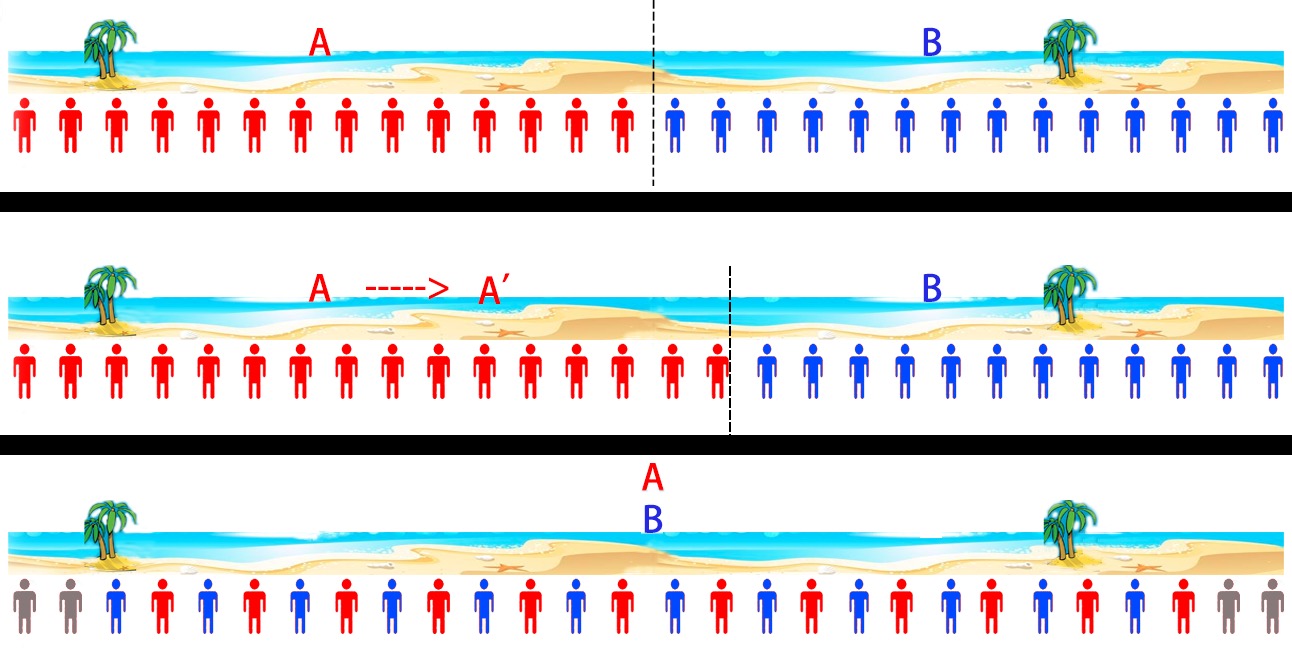}
   %\vspace{-0.1in}
   \caption{A competitive game for two sellers (A and B) selling snacks in a long beach. The top figure shows the initial location, the middle one shows the competing process, and the bottom one shows a solution when the two sellers are competitors. People in red are likely to buy snacks at A, and people in blue at B. People in grey are those beyond the scope of A and B.}
   \label{fig:beach}
\end{figure}

However, ranking strategy in one scenario only optimizes its own metric, without considering the correlation between scenarios (or applications).
%the current situation is that each sub-application only optimize itself and concentrate on the metrics of itself. 
In these platforms, strategies in different scenarios may be developed by different teams, and optimized by different methods with different metrics. Such metrics may include Click Through Rate (CTR), Conversion Rate (CVR), and Gross Merchandise Volume (GMV). However, separate optimization of single scenario cannot guarantee the globally optimal performance of the entire platform. Instead, if the strategies in different scenarios can work collaboratively, we can expect a better overall performance. 
Let's illustrate this with a toy example. In a long beach, as shown in Figure \ref{fig:beach}, there are two sellers (denoted by $A$ and $B$), located at different positions for selling their snacks. The top figure indicates the initial location, where people on the left side of the beach buy snacks at A and people on the right at B. The middle figure shows that when A moves right, he can sell more snacks (A can cover more people than B). Similar cases to B. The bottom figure indicates an optimal solution to this non-cooperative game, where the two sellers compete with each other and they are both at the center of the beach. However, this is a definitely sub-optimal solution if we want to optimize the total income of the two sellers, as some people(in grey) are beyond the scope of them. 

This simple example demonstrates that collaboration between scenarios in a system is extremely important if the objective is to optimize the total return of the system. This is also the case for E-commerce platforms which have many different scenarios in service. In a large E-commerce platform, we indeed observed competitor behaviors: increasing CTR in product search drops that in search advertisement systems, and increasing GMV in main search (the entrance search service of the system) may drop that in-shop search (the search service within a certain shop).
The famous Cournot model~\cite{davidson1986long} can be another example, denoting that if there are more than one oligarch in the market, the total revenue becomes more if the oligarchs are cooperative with each other, but less if they are competitive. 
When ranking strategies in different scenarios are optimized independently but not collaboratively, each strategy maximizes its own objective but ignores the goals of other strategies, leading to a sub-optimal overall performance. We term this issue as the {\bf lack of collaboration between scenarios}. 

Another limitation caused by independent optimization exists in the {\bf inability of modeling the correlation between scenarios}. The user behaviors in different scenarios are correlated and indicative of what they are looking for, which is valuable for optimizing ranking algorithms. Our investigation, on a corpus which consists of user logs of millions of users from Taobao(a large E-commerce platform in China), shows that $25.46\%$ switches from main search to in-shop search and $9.12\%$ switches from in-shop search to main search. 
In addition, scenario switch not only happens between main search and in-shop search, but also among other scenarios such as main search, advertising, and recommendation. Undoubtedly, 
independent optimization in one scenario only uses the partial information (the data within its own scenario) of the user behavior data, which may lead to suboptimal performance. 

In order to deal with the above limitations, we propose a novel model for joint multi-scenario ranking in this paper. The model jointly optimizes ranking strategies for different scenarios through collaboration. In detail, different ranking strategies in a system share an identical goal. The ranking results in one scenario are based on the previous ranking results and user behaviours from all other scenarios. In this way, the ranking strategies collaborate with each other by sharing the same goal; and since each strategy has access to all historical user data across different scenarios, the algorithm within a scenario can make full use of the complete user context.

We cast the multi-scenario ranking task as a fully cooperative, partially observable, multi-agent sequential decision problem.  
The sequential process works as follows: a user enters a scenario, and browses, clicks or buys some items, and then the search system (the model) changes its ranking strategy by adjusting the ranking algorithm when the user navigates into a new scenario or issues a new request. The process is repeated until the user leaves the system. Thus, the current ranking decision definitely affects the following decisions. 

We propose a novel model named Multi-Agent Recurrent Deterministic Policy Gradient (MA-RDPG).
Each ranking strategy in one scenario is treated as an agent. Each agent takes local observations (user behavior data) and makes local actions for ranking items with its private actor network. Different agents share a global critic network to enable them to accomplish the same goal collaboratively. The critic network evaluates the future overall rewards starting from a current state and taking actions. The agents communicate with each other by sending messages. The messages encode historical observations and actions by a recurrent neural network such that agents have access to all historical information. In this manner, our model can optimize ranking strategies in multiple scenarios jointly and collaboratively, and utilizes the complete user behavior data across different scenarios.

The contributions of this paper include:
\begin{itemize}
    \item We formulate multi-scenario ranking (or optimization) as a fully cooperative, partially observable, multi-agent sequential decision problem. 
    
    \item We propose a novel, general multi-agent reinforcement learning model named Multi-Agent Recurrent Deterministic Policy Gradient. The model enables multiple agents (each corresponding to a scenario) to work collaboratively to optimize the overall performance.
    
    \item We evaluate the model with online settings in Taobao, a large online E-commerce platform in China. Results show our model has advantages over strong baselines (Learning-to-rank models).
\end{itemize}

\section{Background}
%To better illustrate our methods, we first introduce some background knowledge.

\subsection{Ranking Strategy} %% 为什么要提这个？
Learning to rank (L2R) \cite{liu2009learning} has been widely applied to deploy ranking strategies in many online platforms. The basic idea of L2R models is that the ranking strategy can be learned and optimized using a set of training samples. Each sample consists of a query and a ranked list of items/documents relevant to that query. The ranking function computes a score for each item with a set of features. The parameters of the ranking function can be learned by various algorithms, such as point-wise~\cite{gey1994inferring,li2008mcrank}, pair-wise~\cite{burges2005learning,qin2007ranking}, and list-wise methods~\cite{burges2007learning,cao2007learning}.

\subsection{Reinforcement Learning}
Reinforcement learning\cite{sutton1998reinforcement} is a framework that enables an agent to learn through interactions with the environment.
At each step $t$, an agent receives the observation $o_t$ of the environment, and takes an action $a_t$ based on a policy $\mu$. The environment changes its state $s_t$, and sends a reward $r_t$ to the agent. The goal of the agent is to find a policy that maximizes the expected cumulative discounted reward $R(s_t,a_t) = r(s_t,a_t) + \gamma r(s_{t+1},a_{t+1}) + \gamma^2 r(s_{t+2},a_{t+2})  + \dots$, where $\gamma$ is a discount factor. 
Generally speaking, reinforcement learning methods can be classified into several branches, including policy-based \cite{sutton2000policy}, value-based \cite{mnih2015human}, and actor-critic \cite{konda2000actor} which combines the two. 

Next, we will give a brief introduction to DDPG~\cite{lillicrap2015continuous} and DRQN~\cite{hausknecht2015deep} models, which are closely related to our proposed model.

\subsubsection{DDPG}
 Deep Deterministic Policy Gradient (DDPG) is an actor-critic approach, which can be applied to solve the continuous action problems.
DDPG maintains a policy function $\mu(s_t)$ and an action-value function $Q(s_t, a_t)$, which are approximated by two deep neural networks respectively, actor network and critic network. The actor network $\mu(s_t)$ deterministically maps a state to a specific action: $a_t=\mu(s_t)$. The critic network $Q(s_t, a_t)$ estimates the future cumulative rewards after taking action $a_t$ at state $s_t$.
In this paper, we employ a deterministic policy where the actor network outputs $a_t=\mu(s_t)$ which corresponds to the weight of a particular feature in a ranking algorithm. In other words, the action in our model is continuous, and DDPG is thus applicable.

\subsubsection{DRQN}
In real-world applications, the state of the environment may be partially observed. The agent is unable to observe the full state of the environment. Such a setting is called partially observable.
Deep Recurrent Q-Networks (DRQN) are introduced to address the partial observation problem by considering the previous context with a recurrent structure. DRQN uses a Recurrent Neutral Network architecture to encode previous observations before the current timestep. Instead of estimating the state-action value function $Q(s_t, a_t)$ in Deep Q-Networks~\cite{mnih2015human}, DQRN estimates $Q(h_{t-1}, o_t, a_t)$, where $h_{t-1}$ is the hidden state of the RNN which encodes the information of previous observations $o_1, o_2, \dots, o_{t-1}$. The recurrent network essentially applies this function to update its hidden states: $h_t=g(h_{t-1},o_t)$ where $g$ is a non-linear function.

\subsection{Multi-Agent Reinforcement Learning}
In multi-agent reinforcement learning (MARL) problems~\cite{busoniu2008comprehensive,littman1994markov,hu1998multiagent,panait2005cooperative}, there are a group of autonomous, interacting agents sharing a common environment. Each agent receives their individual observations and rewards when taking an action based on each individual policy function.
The agents can be fully cooperative, fully competitive, or with mixed strategies. Fully cooperative agents share a common goal and maximize the same expected return. Fully competitive agents have private goals opposite to each other (for instance, zero-sum games). Mixed strategies are in between the two extremes.

\section{Method}
To alleviate the two issues mentioned in the introduction section, we jointly optimize the ranking algorithms in multiple scenarios to maximize the overall returns by casting the task as a multi-agent reinforcement learning problem. We propose a novel model, named Multi-Agent Recurrent Deterministic Policy Gradient (MA-RDPG). In this model, a ranking strategy in one scenario corresponds to an agent, and agents collaborate with each other to accomplish the same goal that optimizes the overall performance.

\subsection{Problem Description}
We formulate this task as a fully cooperative, partially observable, multi-agent sequential decision problem. More specifically:
\\
\textbf{Multi-Agent:} there exist multiple ranking strategies/algorithms for different scenarios in a system. Each agent represents a ranking strategy and learns its own policy function which maps a state to a specific action. 
\\
\textbf{Sequential Decision:} users sequentially interact with the system. Thus, the agent actions are also sequential. At each step, the agent, which represents the scenario interacting currently with the users, chooses an action to respond to the user through a sorted list of items. The current actions affect the following actions in the future.
\\
\textbf{Fully Cooperative:} all agents are fully cooperative to maximize a shared metric. Moreover, the agents pass messages to each other for communication, and the overall performance of these agents are evaluated by a centralized critic.  
\\
\textbf{Partially Observable:} 
The environment is partially observable, and each agent only receives a local observation instead of observing the full state of the environment.
\begin{figure}[t]
 \centering
 \includegraphics[width=0.5\textwidth]{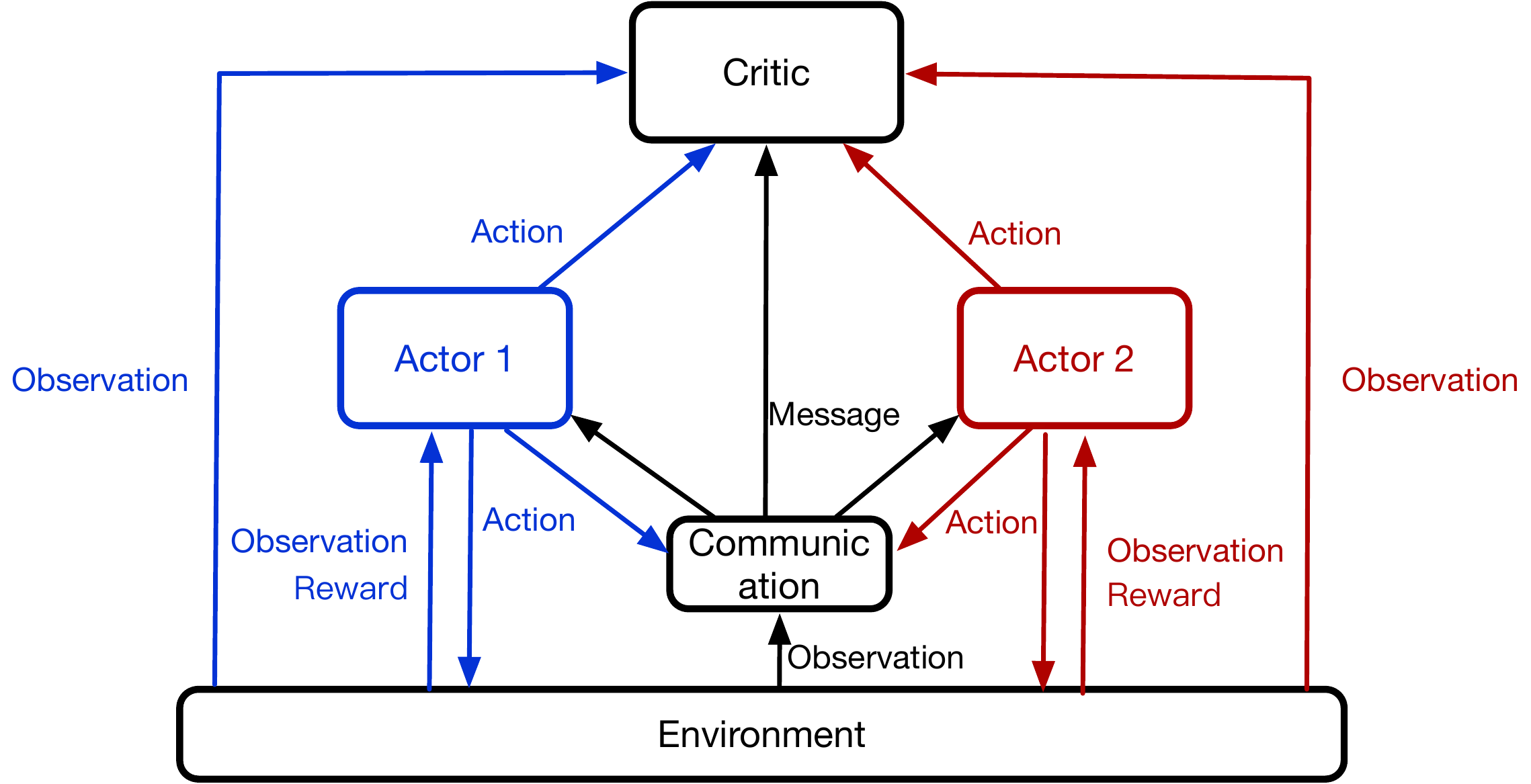}
 \caption{Overall model architecture. The model has a centralized, global critic network to evaluate the overall rewards. A communication module is used to generate messages that are shared among actors. Messages encode historical observations and actions, and can be used to approximate the global state of the environment. Each actor network represents an agent which receives its own local observations and a communication message, and makes private actions.  }
 \label{fig:framework}
\end{figure}

\subsection{Model}
We design a Multi-Agent Recurrent Deterministic Policy Gradient~(MA-RDPG) model to address the fully cooperative, partially observable, multi-agent sequential decision problem. 

\subsubsection{Overview}
Figure~\ref{fig:framework} shows the overall architecture of our model.
For simplicity, we consider the case with two agents, each agent representing a scenario or strategy to be optimized.
Inspired by DDPG~\cite{lillicrap2015continuous}, our model is built on top of the actor-critic approach \cite{konda2000actor}. We design three key modules to enable the agents to collaborate with each other: a centralized critic, private actors, and a communication component.
The centralized critic evaluates an action-value function that indicates the expected future rewards for all agents taking actions from the current state. Each agent is represented by an actor network which maps a state to a specific action with a deterministic policy. Actions made by each actor network will be used for the agent to perform optimization in its own scenario.

\begin{figure}[t]
 \centering
 \includegraphics[width=0.45\textwidth]{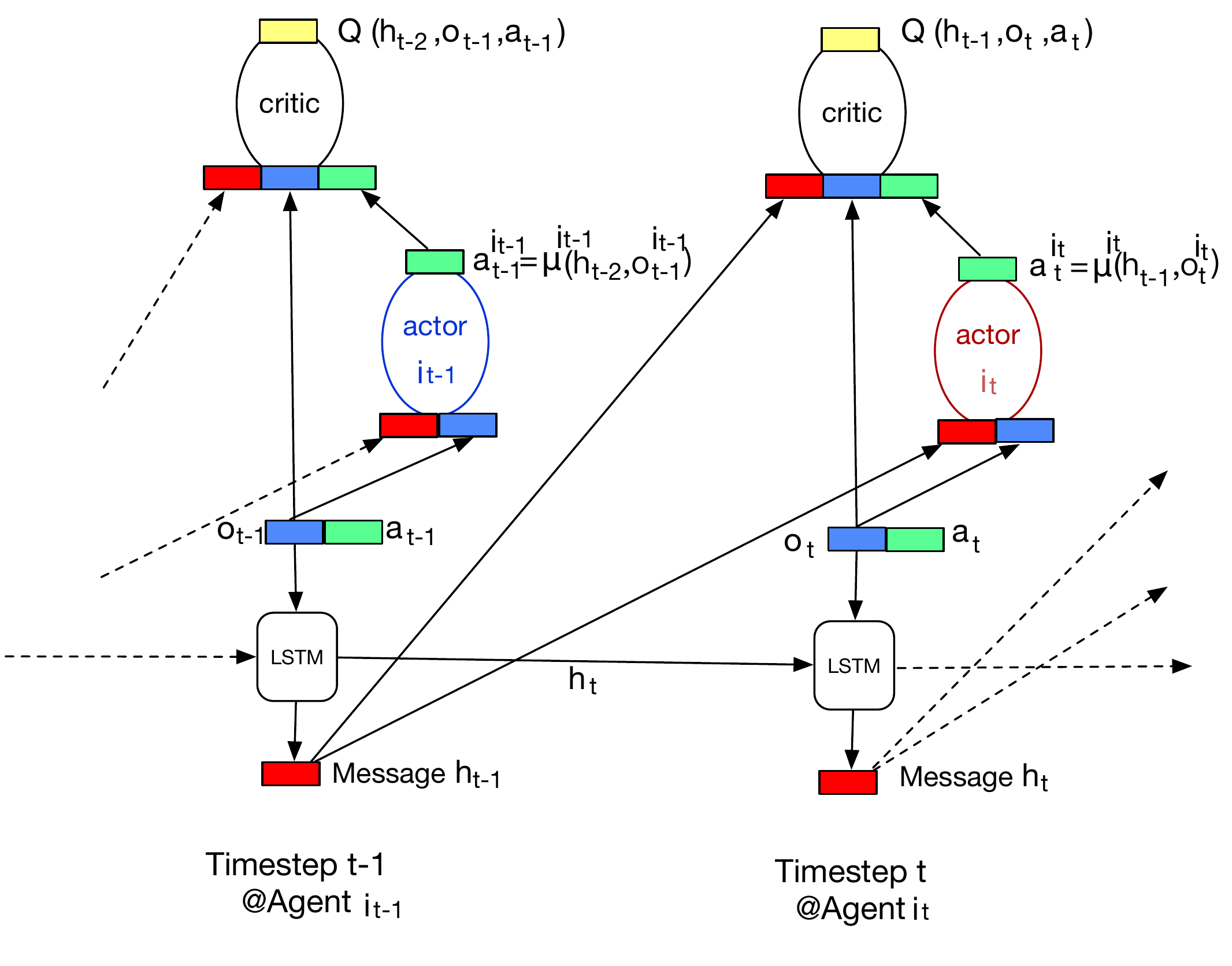}
 \caption{Detailed structure of MA-RDPG. The centralized critic network estimates the action-value function $Q(h_{t-1},o_t,a_t)$ which indicates the future overall rewards when taking action $a_t$ upon observing message $h_{t-1}$ and observation $o_t$. The actor network outputs a deterministic action with $a^i_t=\mu^i(h_{t-1},o^i_t)$ given the message and local observation as input. The messages are updated by a communication component which takes as input the observation $o_t$ and action $a_t$. Red: Message; Blue: Observation; Green: Action. }
 \label{fig:model}
\end{figure}

We design a communication component using a Long Short-Term Memory (LSTM) architecture~\cite{hochreiter1997long}. The LSTM encodes all local observations and the actions of all agents into a message vector. The message will be sent between agents for collaboration. Thanks to this component, the decision of each agent depends not only on its own previous observations and actions, but also on other agents' observations and actions. In addition, the messages can help the agents approximate the full state of the environment, which enables them to act more efficiently.

\subsubsection{Model Details}
A general reinforcement learning problem has a sequence of experiences $(o_1, r_1, a_1, \cdots, a_{t-1}, o_t, r_t)$ where $o/r/a$ correspond to observation/reward/action respectively. As aforementioned, the environment in our problem is partially observable. In other words, the state $s_t$ is the summary of the previous experiences: $s_t=f(o_1, r_1, a_1, \cdots, a_{t-1}, o_t, r_t)$\footnote{In a fully observable environment, $s_t=f(o_t)$.}. We are considering the problem with $N$ agents $\{A^1, A^2, \dots, A^N\}$, each agent corresponding to a particular optimization scenario (ranking, recommendation, etc.). In this multi-agent setting, the state of the environment ($s_t$) is global, shared by all agents, while the observation ($o_t=(o^1_t,o^2_t,\cdots,o^N_t)$), the action ($a_t=(a^1_t,a^2_t,\cdots,a^N_t)$), and the intermediate reward ($r_t=(r(s_t,a^1_t),r(s_t,a^2_t), \cdots, r(s_t,a^N_t)$)) are all private, only possessed by each agent itself. 

More specifically, each agent $A^i$ takes action $a^i_t$ with its own policy specified by $\mu^i(s_t)$, and obtains a reward $r^i_t = r(s_t, a^i_t)$ from the environment which changes its current state $s_t$ to the next state $s_{t+1}$. 
In our task, all agents are collaborating to achieve the same goal. 
This leads to a collaborative setting of multi-agent reinforcement learning. We have a centralized action-value function $Q(s_t,a_t^1,a_t^2,\cdots,a_t^N)$ (as critic) to evaluate the future overall return when taking the actions ($a_t^1,a_t^2,\cdots,a_t^N$) at the current state. We also have a global state representation of the environment, and each agent is represented by a private actor which observes local observations and takes private actions. Thus, the model belongs to an actor-critic reinforcement learning approach with a centralized critic and several private actors (each actor plays its role as an agent).

As shown in Figure~\ref{fig:model}, at step $t$, agent $A^{i_t}$ receives an current local observation $o^{i_t}_t$ from the environment. The global state of the environment, shared by all agents, depends not only on all the historical states and actions of all agents in the sequential decision process, but also the current observation $o_t$. In other words, $s_t=f(o_1,a_1,\cdots,a_{t-1},o_t)$\footnote{Intermediate rewards $r_t$ can be omitted in general for state representation.}. To this end, we design a communication component using LSTM to encode the previous observations and actions of all agents into a message vector. With the message $h_{t-1}$ sent between agents, the full state can be approximated as $s_t \approx \{h_{t-1}, o_t\}$ since the message $h_{t-1}$ has encoded all previous observations and actions (see soon later). Agent $A^{i_t}$ chooses the action $a^{i_t}_t = \mu^{i_t}(s_t)\approx \mu^{i_t}(h_{t-1}, o^{i_t}_t)$ with the purpose of maximizing the future overall rewards estimated by the centralized critic $Q(s_t, a^1_t,a^2_t,\cdots,a^N_t)$. Note that at each timestep, $o_t=(o^1_t,o^2_t,\cdots,o^N_t)$ consisting of observations by all agents.

\begin{figure}[t]
 \centering
  \includegraphics[width=0.45\textwidth]{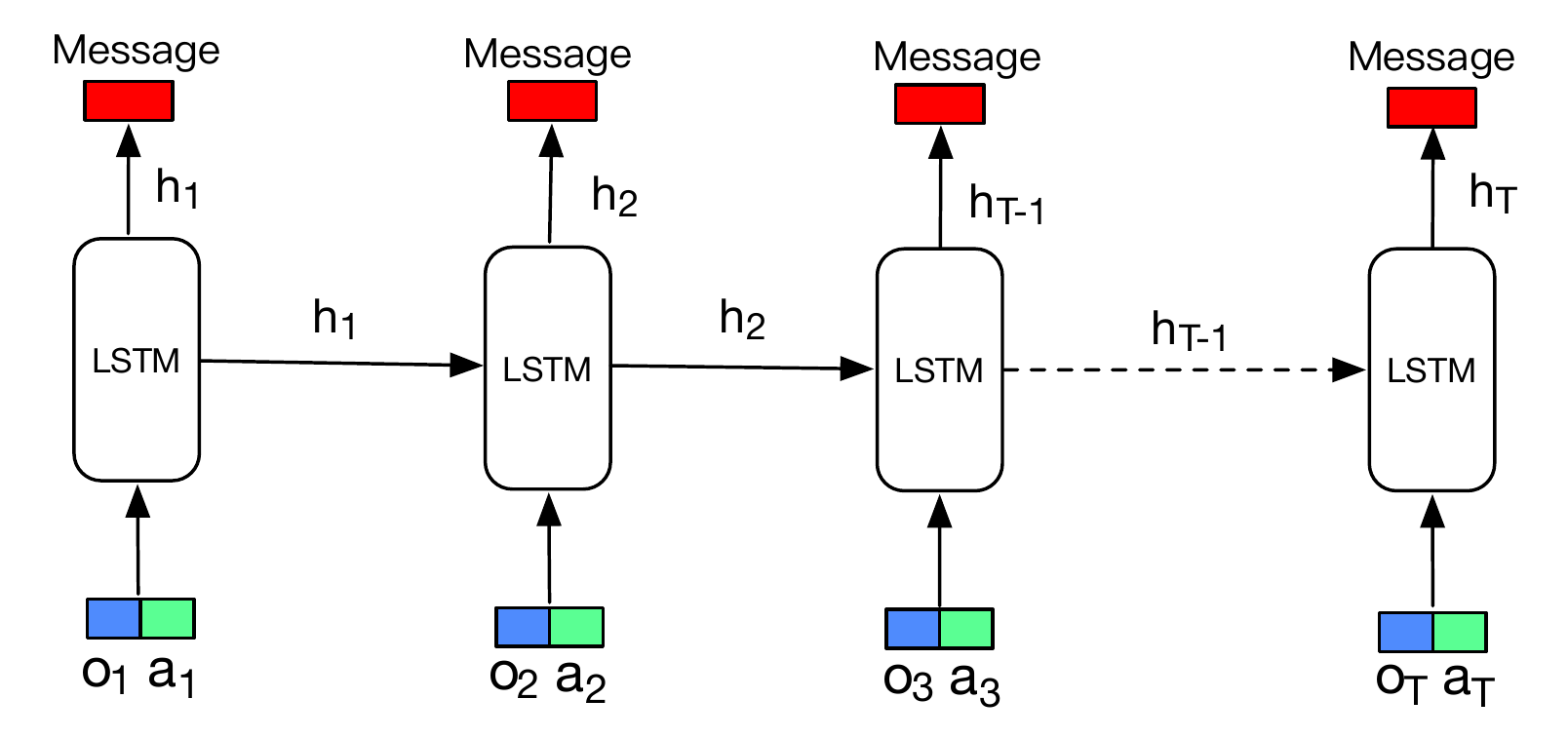}
  %\vspace{-0.1in}
  \caption{Communication component. The previous observations ($o_t$) and actions ($a_t$) are all taken as input to the LSTM network. The hidden states ($h_{t-1}$) are treated as messages which will be sent between agents. Note that $o_t,a_t$ are vectors.}
  \label{fig:message}
 %\vspace{-0.1in}	
 \end{figure}

\textbf{Communication Component}
We design a communication component to make the agents collaborate better with each other by sending messages. The message encodes the local observation and the actions at previous steps. At step $t$, agent $A^{i_t}$ receives an local observation $o^{i_t}_t$ and a message $h_{t-1}$ from the environment.
The communication component generates a new message $h_t$  taking as input the previous message $h_{t-1}$ and current observation $o_t$. An agent can share the information with other collaborators through the message.
As shown in Figure~\ref{fig:message}, we apply a LSTM architecture for this purpose. Formally, the communication component works as follows:
\begin{equation}
    h_{t-1}=LSTM(h_{t-2},[o_{t-1};a_{t-1}]; \psi)
\end{equation} %%注意这里要用t-1时刻，更清楚
Note that $o_t$ and $a_t$ consists of observations and actions of all agents respectively, and each action $a^i_{t}$ is also a real-valued vector since our problem is a continuous action reinforcement learning problem. 
%The input of the LSTM are the $h_{t-1}$ and $o_t$, and the output is the new message $h_t$. 

With the help of the message $h_{t-1}$, agents have access to an approximate of the full state of the environment: $s_t \approx \{h_{t-1}, o_t\}$, as an agent only receives its current observation $o^i_t$ but not the full state $s_t$ of the environment. 
%The state $s_t = \{s_1, a_1, \dots, s_{t-1}, a_{t-1}, o_t\}$ includes the historical states and actions in the sequential decision process and the current local observation $o_t$.

\textbf{Private Actor}
Each agent has a private actor which receives local observations and shared messages, and makes its own actions. Since we deal with continuous action problems, we define the agent's action as a vector of real values, $a^i = (w^i_1, \dots, w^i_{N^i}), a^i \in \mathbb{R}^{N^i}$. Therefore, an action is a $N^i$-dimension vector, and each dimension is a continuous value. The action vector will be used in ranking algorithms or to control robots.

Since this is a continuous action problem which can be commonly seen in control problems ~\cite{santamaria1997experiments,lillicrap2015continuous,heess2015learning}, we resort to using a deterministic policy instead of a stochastic policy.
The actor of each agent $\mu^i(s_t;\theta^i)$, parameterized by $\theta^i$, specifies a deterministic policy that maps states to a specific action. At timestep $t$, agent $A^{i_t}$ takes an action with its own actor network:
\begin{equation}
a_t^{i_t} = \mu^{i_t}(s_t;\theta^{i_t}) \approx \mu^{i_t}(h_{t-1}, o^{i_t}_t;\theta^{i_t})
\end{equation}
where $s_t\approx\{h_{t-1},o_t\}$ as discussed in the communication component. In this manner, the actor is conditioned on the message $h_{t-1}$ and its own current local observation $o^{i_t}_t$. 
%The action is the feature weight of the learning-to-ranking model.

\textbf{Centralized Critic}
Following DDPG, we design a critic network estimating the action-value function to approximate the expected future total rewards.
As all agents share the same goal, we use a centralized critic $Q(s_t, a^1_t, a^2_t, \cdots, a^N_t;\phi)$ to estimate the future overall rewards obtained by all agents after taking action $a_t = \{a_t^1, \dots, a_t^N \}$ at state $s_t \approx \{h_{t-1}, o_t\}$. 

The above formulation is general and applicable to many agents that are alive all the time. In our setting\footnote{Because a user can be in only one physical scenario at each timestep.}, there is only one agent $A^{i_t}$ activated at timestep $t$, and $o_t = \{o_t^{i_t}\}$ and $a_t = \{a_t^{i_t}\}$. Hereafter, we will simplify the action-value function as $Q(h_{t-1}, o_t, a_t;\phi)$ and policy function as $\mu^{i_t}(h_{t-1}, o_t; \theta^{i_t})$.
\begin{algorithm}[!t]
Initialize the parameters $\theta = \{\theta^1, \dots, \theta^N\}$ for the $N$ actor networks and $\phi$ for the centralized critic network. \\
% Initialized the target network $\theta' = \theta, \phi' = \phi$ \\
Initialized the replay buffer $R$ \\
\For {each training step $e$}
{
    \For {i = 1 to M}
    {
        $h_0$ = initial message, $t = 1$ \\
        \While {$t < T$ and $o_t \neq terminal$}
        {
            Select the action $a_t = \mu^{i_t}(h_{t-1}, o_t)$ for the active agent $i_t$ \\
            Receive reward $r_t$ and the new observation $o_{t+1}$\\
            Generate the message $h_t = LSTM(h_{t-1}, [o_t; a_t])$
            $t = t + 1$
        }
        %\EndWhile
        Store episode $\{h_0, o_1, a_1, r_1, h_1, o_2, r_2, h3, o3, \dots\}$ in $R$
    }
    Sample a random minibatch of episodes $B$ from replay buffer $R$ \\
	\ForEach {episode in $B$}
	{
	    \For {t = T downto 1}
	    {
	        Update the critic by minimizing the loss: $L(\phi) = (Q(h_{t-1}, o_t, a_t; \phi) - y_t)^2$, where $y_t = r_t + \gamma Q(h_t, o_{t+1}, \mu^{i_{t+1}}(h_{t}, o_{t+1}); \phi)$\\
	        Update the $i_t$-th actor by maximizing the critic:
	        $J(\theta^{i_t}) = Q(h_{t-1}, o_t, a; \phi) |_{a = \mu^{i_t}(h_{t-1}, o_t;\theta^{i_t})}$ \\
	        Update the communication component.
	    }
	}
% 	Update the weights of the target networks:\\
% 	    \quad $\theta' = \tau \theta + (1 - \tau) \theta'$ \\
% 	    \quad $\phi' = \tau \phi + (1 - \tau) \phi'$
}
\caption{MA-RDPG}
\label{alg:paper}
\end{algorithm}

\subsection{Training}
The centralized critic $Q(h_{t-1}, o_t, a_t; \phi)$ is trained using the Bellman equation as in Q-learning~\cite{watkins1992q}. We minimize the below loss:
\begin{equation}
L(\phi) = \mathbb{E}_{h_{t-1}, o_t}[(Q(h_{t-1}, o_t, a_t; \phi) - y_t)^2]
\end{equation}
where
\begin{equation}
y_t = r_t + \gamma Q(h_t, o_{t+1}, \mu^{i_{t+1}}(h_{t}, o_{t+1}); \phi)
\end{equation}

The private actor is updated by maximizing the expected total rewards with respect to the actor's parameters. If agent $A^{i_t}$ is active at step $t$, the objective function is:
\begin{equation}
J(\theta^{i_t}) = \mathbb{E}_{h_{t-1}, o_t} [ Q(h_{t-1}, o_t, a; \phi) |_{a = \mu^{i_t}(h_{t-1}, o_t;\theta^{i_t})}]
\end{equation}
Following the chain rule, the gradients of the actor's parameters are given as below:
\begin{equation}
\begin{aligned}
& \nabla_{\theta^{i_t}} J(\theta^{i_t}) \\ & \approx \mathbb{E}_{h_{t-1}. o_t}[\nabla_{\theta^{i_t}} Q(h_{t-1}, o_t, a; \phi) |_{a = \mu^{i_t}(h_{t-1}, o_t;\theta^{i_t})}] \\ & = \mathbb{E}_{h_{t-1}, o_t} [\nabla_{a} Q(h_{t-1}, o_t, a; \phi) |_{a=\mu^{i_t}(h_{t-1}, o_t)} \nabla_{\theta^{i_t}} \mu^{i_t}(h_{t-1}, o_t;\theta^{i_t})]
\end{aligned}
\end{equation}

The communication component is trained by minimizing:
\begin{equation}
\begin{aligned}
    & L(\psi) \\ & = \mathbb{E}_{h_{t-1}, o_t}[(Q(h_{t-1}, o_t, a_t; \phi) - y_t)^2|_{h_{t-1} = LSTM(h_{t-2}, [o_{t-1};a_{t-1}];\psi)}] \\ & - \mathbb{E}_{h_{t-1}, o_t} [ Q(h_{t-1}, o_t, a_t; \phi) |_{h_{t-1} = LSTM(h_{t-2}, [o_{t-1};a_{t-1}];\psi)}]
\end{aligned}
\end{equation}

The training process is shown in Algorithm ~\ref{alg:paper}. 
We use a replay buffer~\cite{lillicrap2015continuous} to store the complete trajectories to learn with minibatch update, rather than online update. At each training step, we sample an minibatch of  episodes and process them in parallel to update the actor networks and the critic network respectively.
% In order to have a stable update, we take advantage target network. We set $N$ target actors and a target critic with parameter sets $\theta' = \{\theta^1, \dots, \theta^N\}$ and $\phi'$ respectively, similar to ~\cite{lillicrap2015continuous}. The parameters in the target networks are updated much more slowly than the original ones. We update $\theta'$ and $\phi'$ by linear interpolation: $\theta' \leftarrow (1 - \tau) \theta' + \tau \theta$ and $\phi' \leftarrow (1 - \tau) \phi' + \tau \phi$, where $\tau \ll 1$ is a hyper-parameter. 
%sharing parameters

\section{Application}
Previous sections present a general multi-agent reinforcement learning framework that may be applicable to many joint optimization scenarios.
To evaluate the proposed model, we apply it to jointly optimize the ranking strategies in two search scenarios in Taobao, which is a real-world E-commerce platform.

\begin{figure}[htbp]
	\centering
	\subfigure[The two search engines are optimized separately.]{
		\begin{minipage}[b]{0.45\textwidth}
		    \centering
			\includegraphics[width=0.9\textwidth]{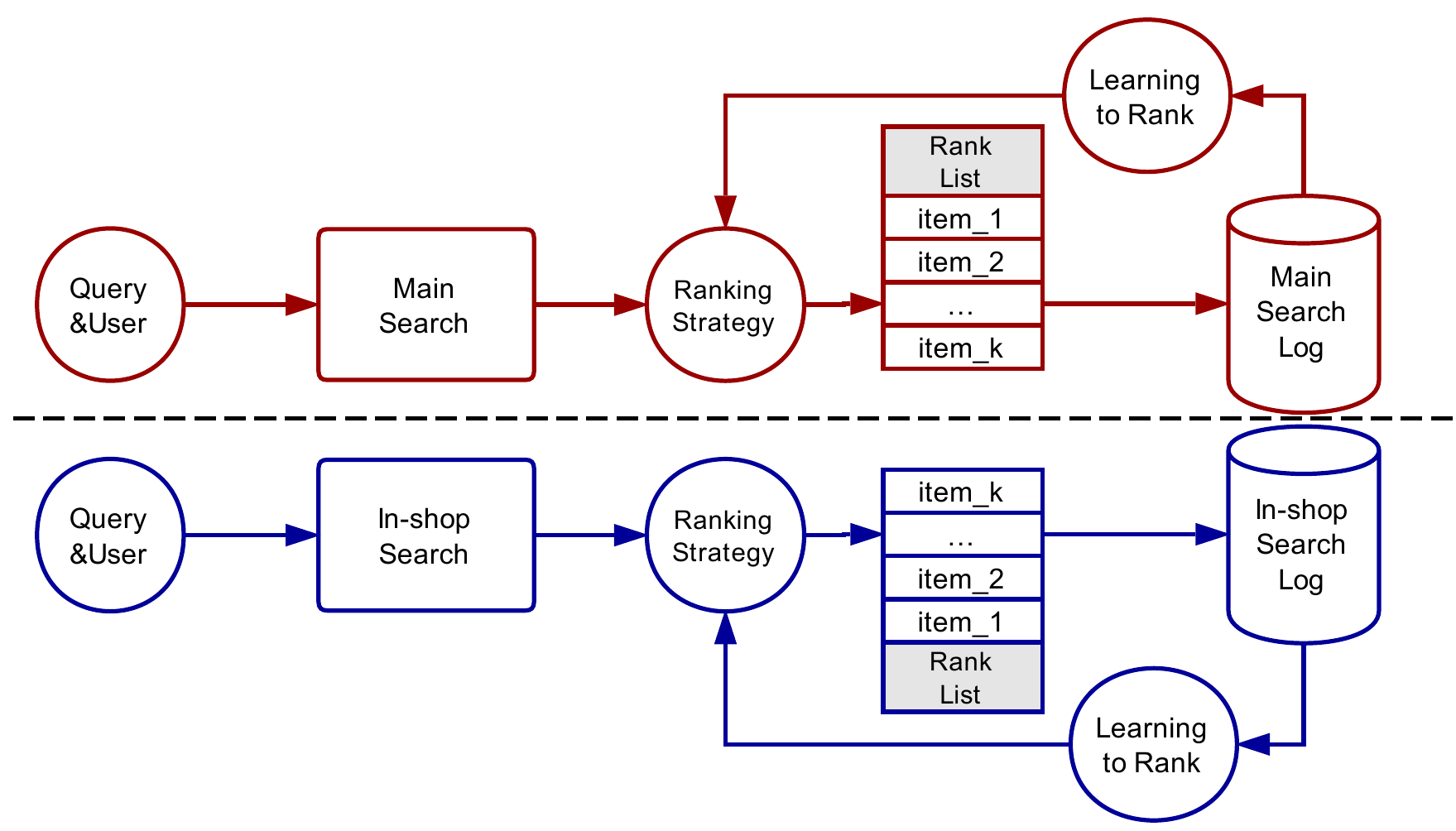} 
		\end{minipage}
	}
	\subfigure[The two systems work collaboratively in MA-RDPG.]{
		\begin{minipage}[b]{0.45\textwidth}
		    \centering
			\includegraphics[width=0.9\textwidth]{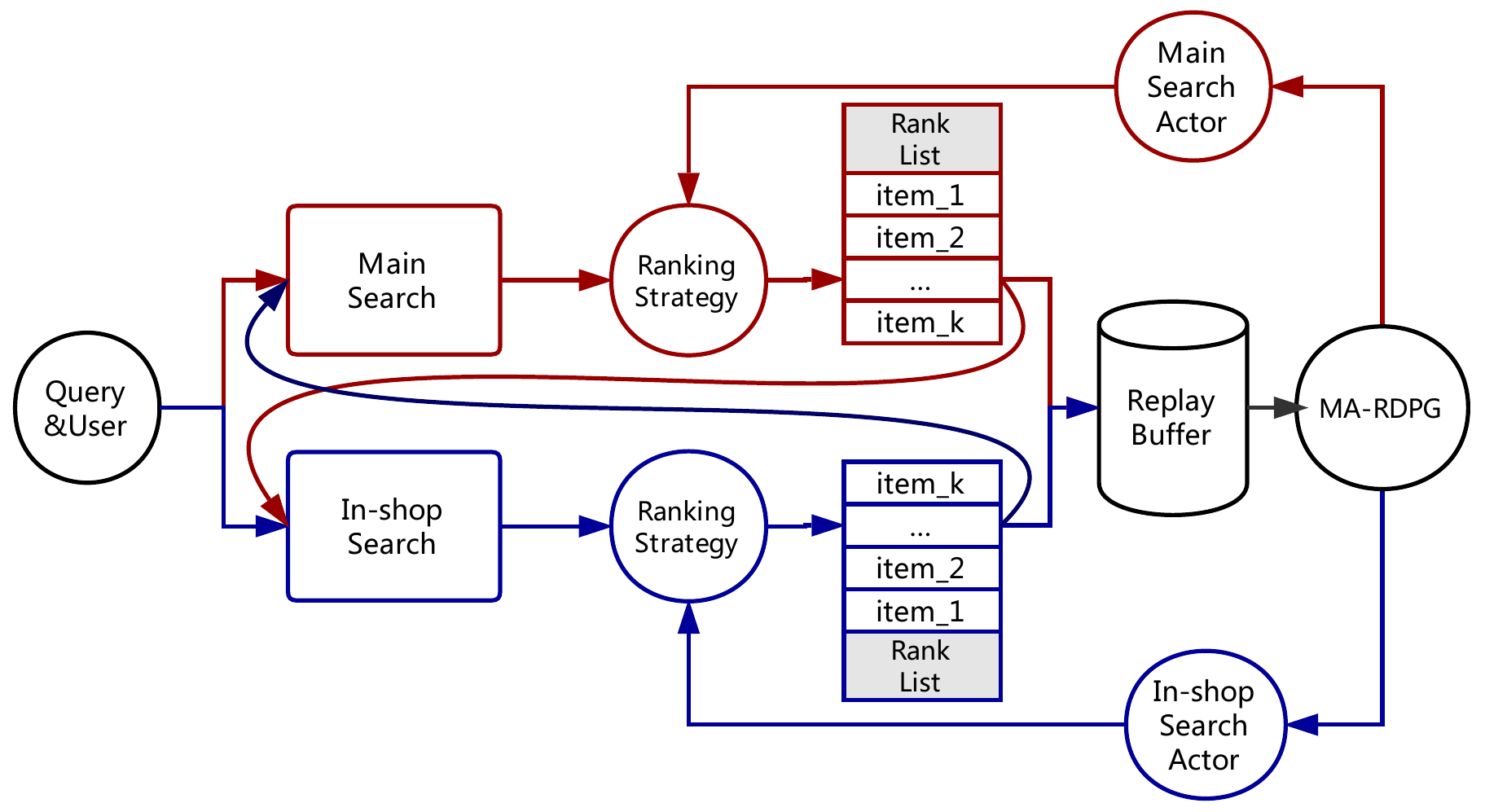} 
		\end{minipage}
	}
	%% Marl --> MARL; In-shop Actor; Main Search Actor 这样的
	\caption{Comparison of two search systems that are optimized separately or collaboratively. %The left figure illustrates the case that the two search engines are optimized independently. In the right figure, the search scenarios are believed to correlate with each other and ranking strategies are optimized jointly and collaboratively.
	 }                     
	\label{fig:system}   
\end{figure}

Firstly, we give a brief overview of the online E-commerce platform. Then, we explain the details of how we apply our MA-RDPG to Taobao.

\subsection{Search Scenarios of an E-commerce Platform}
An E-commerce platform generally consists of multiple search scenarios, each of which has its own ranking strategy. In particular, we choose two important search scenarios of an E-commerce platform for this study: the main search and the in-shop search. The two search types are detailed as follows:
%The search advertising ranks advertisements accompany with normal search products. And various kinds of recommenders rank items based on users' interest and behaving history.

\textbf{Main search} ranks the relevant items when a user issues a query through the search box in the entrance page of the E-commerce platform. The main search returns various items from different sub-domains in the platform. The main search occupies the majority of the user traffic. In our platform, there are about 40,000 queries of main search per second. Within one day, there could be about 3.5 billion page views and 1.5 billion clicks from more than 100 million customers.

\textbf{In-shop search} ranks items in a certain shop when a user browses products at a shop's page \footnote{Some E-commerce systems such as Taobao or JingDong are the same as the real marketplaces which have many shops. Each shop sales its own products.}. During the in-shop search, customers can search either with an input query or without any query. In one day, more than 50 million customers make shopping via in-shop search, amounting to 600 million clicks and 1.5 billion page views.

Users constantly navigate cross the two scenarios.
When an user find a dress that she likes in the main search, she may go into the shop site for more similar products. When the user finds that the clothes in the shop are too limited, the user may go back to the main search for more products from other shops. Our investigation suggests that among all user shopping behavior data in Taobao, $25.46\%$ switches from the main search to the in-shop search and $9.12\%$ switches from the in-shop search back to the main search.  %%

In existing models \cite{Covington2016Deep, Kenthapadi2017LiJAR, Gupta2015Complementary}, %% reference???
different ranking strategies in different scenarios are independently optimized, and each strategy maximizes its own objective and ignores those of the other strategies. Figure ~\ref{fig:system}(a) describes a traditional optimization method for dealing with multiple search scenarios in online platforms. The upper block in red denotes the main search engine and the lower block in blue denotes the in-shop search engine. The two search engines are optimized separately and independently. 

\subsection{Joint Optimization of Multi-scenario Ranking}
We illustrate a solution to jointly optimizing ranking strategies in the main search and in-shop search in Figure ~\ref{fig:system}(b). Instead of separately optimizing the ranking strategies in the two search scenarios, MA-RDPG employs two agents (actors) to model the two strategies collaboratively. The main search and in-shop search actors learn the weights of features in the ranking algorithms for the two scenarios respectively. The two actors collaborate in two ways: First, they have the same goal to optimize the overall performance of the system; Second, they share and broadcast messages through the communication component such that both of them have access to all historical information in different scenarios.

To be concrete, we will introduce the key concepts when MA-RDPG is applied to the scenarios.

\textbf{Environment.}
The environment is the online E-commerce platform. Its state changes when the two agents (actors) take actions to present different ranking items. It offers rewards to the actors which also take as input the observations from the environment.

\textbf{Agents.} 
There are two agents: one is the search engine for main search and the other is that for in-shop search. At each step, one of the search engines returned a ranked list of products according to the ranking algorithm (linearly summing the features values with the feature weights). The two agents work together to maximize the overall performance, GMV, for instance.

\textbf{States.}
As aforementioned, the states are partially observable. Agents can only receive a local observation which includes: the attributes of the  customer (age, gender, purchasing power, etc.), the properties of the customer's clicked items (price, conversion rate, sales volume, etc.), the query type and the scenario index (main or in-shop search). A 52-dimension vector is then formed to represent the observed information.
%We transform each of these role information into a one-hot vector and combine them into a 50-dimensional local observation vector. 
As shown in MA-RDPG, the complete state vector are concatenation of the local observation vector and message vector which encodes historical observations and actions.

\begin{figure}[t]
 \centering
   \includegraphics[width=0.45\textwidth]{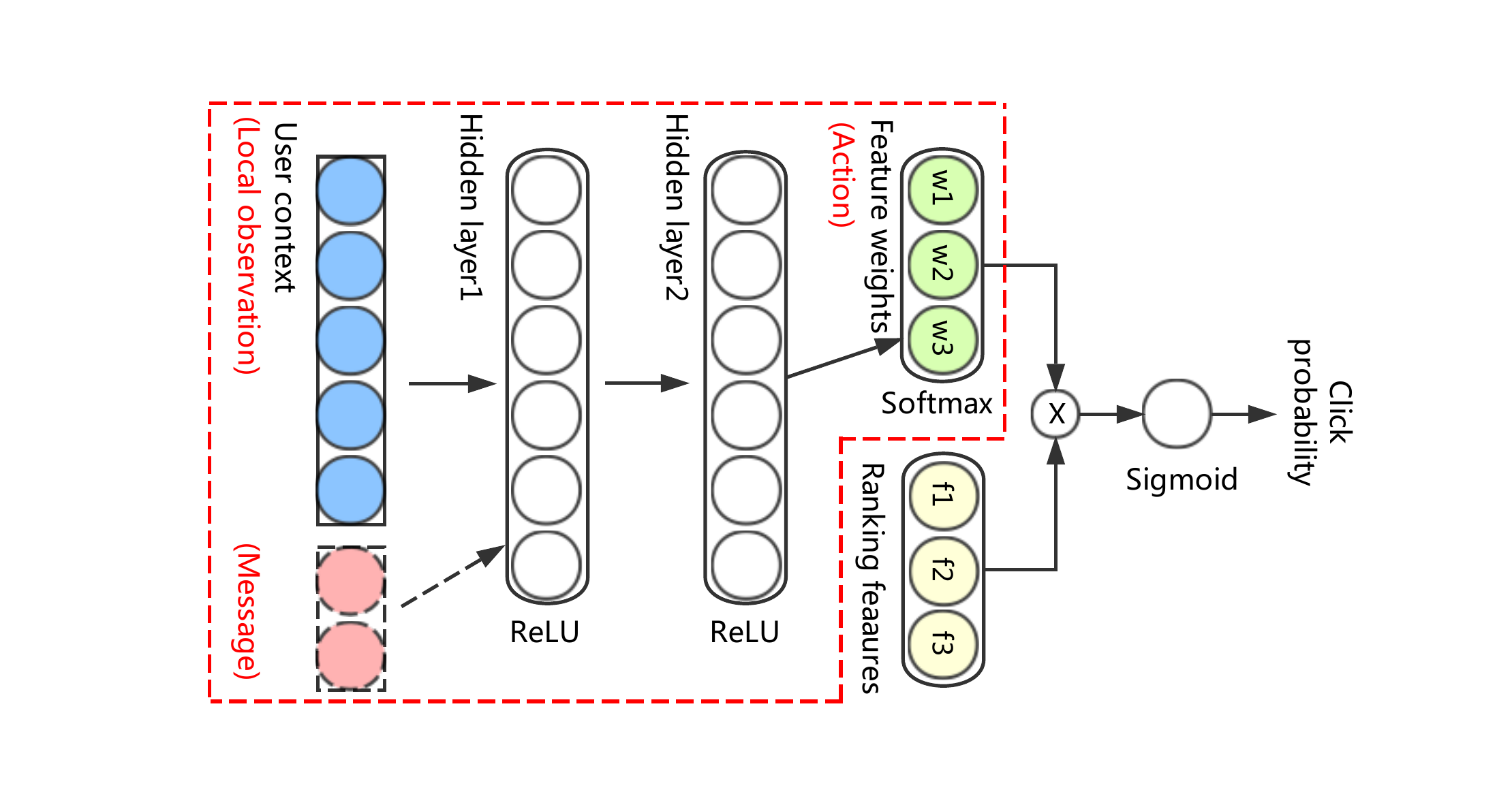}
   %\vspace{-0.1in}
   \caption{Actor network. The actor network in red dashed box outputs an real-valued action vector (green) for ranking given the input of local observation (blue) and message(red). }
   \label{fig:ltr}
 %\vspace{-0.1in}	
\end{figure}

\textbf{Actions.}
The agent needs to provide a ranking list of relevant items in response to an input query (or sometimes no query). Thus, the action of the agents is defined as the weight vector for the ranking features. To rank items, the ranking algorithm computes an inner product of the feature value vector and the weight vector. Changing an action means to change the weight vector for the ranking features. For main search, we set the actor's action as a 7-dimension real-valued vector. For in-shop search,  the action is a 3-dimension real-valued vector. %The features are limited to the current settings of the algorithms running in the platform, but can be generalized to other settings.

Each agent has its own policy function. The architecture of the actor network is shown in Figure \ref{fig:ltr}. The actor network is a three-layer Perceptrons (MLP) with ReLu activation functions for the first two layer and softmaxt for the output layer. The input to the actor network is the local observation vector and the message vector. The output is the weight vector for the ranking features.
%As the action dimension of the two agents are different, we adopt a padding strategy to transform the action into same dimension.

\textbf{Reward.}
%Though the objective we want to optimize is GMV, 
We design the rewards by considering not only purchase behaviors but also other user behaviors. In this manner, we can make full use of user feedback on the presented product list.
%
%The reward is provided by the online platform and related to the customer's behavior after the agent returns a ranking list of products.
 If a purchase behavior happens, there is a positive reward that equals to the price of the bought product. If a click happens, there is a positive reward of $1$. If there is no purchase nor click, a negative reward of $-1$ is received. If a user leaves the page without buying any product, there is a negative reward of $-5$.

\begin{table}[h]
    \small
	\centering
	\caption{Examples of Ranking Features}
	\begin{tabular}{|c|c|c|}
		\hline
		Scenario& Feature Name & Description \\
		\hline
		
		\multirow{4}[-2]*{Main}&\multirow{3}[-6]*{Click}& An CTR estimation using logistic \\     
		
		& Through& regression, considering features of \\
		& Rate& users, items and their interactions \\\cline{2-3}
		
		Search&\multirow{1}[1]*{Rating Score}& Average user ratings on a certain item\\ \cline{2-3} 
		%\hline
		
		&\multirow{1}[1]*{Shop Popularity}& Popularity of the item shop  \\  \hline   
     %% of an item是啥意思	
		
		\multirow{4}[-6]*{In-shop}&\multirow{2}[0]*{Latest Collection }& Whether an item is the latest   \\
		
		&  & collection or new arrivals of the shop \\ \cline{2-3}
		
		Search& \multirow{1}[1]*{Sales Volume} & Sales volume of an in-shop item \\     \cline{2-3}
	
		\hline 		 	 	 	
	\end{tabular}
	
	\label{Feature}
\end{table}

\section{Experiment}
To evaluate the performance of our proposed MA-RDPG model, we deployed our model on Taobao  to jointly optimize the main search and in-shop search.

\subsection{Experiment Setting}
\textbf{Training Process.} The flow chart of our model is shown in Figure \ref{fig:system}(b). Our training process is based on an online learning system which consumes unbounded streams of data. Firstly, the system collects user logs in real time and provides training episodes for MA-RDPG. Secondly, the episodes are stored in a replay buffer. Thirdly, gradients are computed to update model parameters using the episodes sampled from the replay buffer. At last, a new, updated model is deployed to the online system. The process repeats. Thus, the online model is changing periodically and dynamically to capture the dynamics of user behaviors.

\textbf{Parameter Setting.} 
For each agent, the local observations is a 52-dimensional vector. The dimension of the action vector is 7 and 3 for the main and in-shop search respectively. As the communication component and critic network will take the action vectors of both actors as input, for convenience, a vector of a normalized length 10 (7+3) with zero-padding is taken as input to the LSTM and critic networks. %If the action vector is from main search, the first 7 dimension is from the action vector and the last 3 dimension are all equal to $0$.

For the communication component, the input is a $52+7+3=62$ dimensional vector and the output message is a 10-dimension vector. The network structure is shown in Figure \ref{fig:message}.
In the actor network, the dimension of the input layer is $52+7+3=62$. The actor network was parameterized by a three-layer MLP with 32/32/7 (or 3) neurons for the first/second/third layer, respectively. The activation functions are ReLU for the first two layers and softmax for the output layer. The network structure is shown in Figure \ref{fig:ltr}.   
The critic network has two hidden layers with 32 neurons per layer. The ReLU activation function is also used.

The reward discount factor is $\gamma=0.9$. 
In our experiments, we used RMSProp for learning parameters with a learning rate of $10^{-3}$ and ${10^{-5}}$ for the actor and critic network respectively. We used a replay buffer size of $10^{4}$ and the minibatch size is 100. 

\begin{table*}
\centering
\caption{GMV gap evaluated on an online E-commerce platform. A+B means algorithm A is deployed for the main search and B for the in-shop search. The values are the relative growth ratio of GMV compared with the EW+EW setting. }
\begin{tabular}{|*{13}{r|}}
\hline
\multicolumn{1}{|c|}{\multirow{2}*{day}}
& \multicolumn{3}{|c|}{EW + L2R} & \multicolumn{3}{|c|}{L2R + EW}
& \multicolumn{3}{|c|}{L2R + L2R} & \multicolumn{3}{|c|}{MA-RDPG(ours)}\\\cline{2-13}
 & main   & in-shop & total & main & in-shop & total & main & in-shop & total & main & in-shop & total \\\hline
1 & 0.04\% & 1.78\% & 0.58\% & 5.07\% & -1.49\% & 3.04\% & 5.22\% & 0.78\% & 3.84\% & 5.37\% & 2.39\% & 4.45\%\\
2 & 0.01\% & 1.98\% & 0.62\% & 4.96\% & -0.86\% & 3.16\% & 4.82\% & 1.02\% & 3.64\% & 5.54\% & 2.53\% & 4.61\%\\
3 & 0.08\% & 2.11\% & 0.71\% & 4.82\% & -1.39\% & 2.89\% & 5.02\% & 0.89\% & 3.74\% & 5.29\% & 2.83\% & 4.53\%\\
4 & 0.09\% & 1.89\% & 0.64\% & 5.12\% & -1.07\% & 3.20\% & 5.19\% & 0.52\% & 3.74\% & 5.60\% & 2.67\% & 4.69\%\\
5 & -0.08\% & 2.24\% & 0.64\% & 4.88\% & -1.15\% & 3.01\% & 4.77\% & 0.93\% & 3.58\% & 5.29\% & 2.50\% & 4.43\%\\
6 &0.14\% & 2.23\% & 0.79\% & 5.07\% & -0.94\% & 3.21\% & 4.86\% & 0.82\% & 3.61\% & 5.59\% & 2.37\% & 4.59\%\\
7 & -0.06\% & 2.12\% & 0.62\% & 5.21\% & -1.32\% & 3.19\% & 5.14\% & 1.16\% & 3.91\% & 5.30\% & 2.69\% & 4.49\%\\\hline
avg. & 0.03\% & 2.05\% & 0.66\% & 5.02\% & -1.17\% & 3.09\% & 5.00\% & 0.87\% & 3.72\% & 5.43\% & 2.57\% & 4.54\%\\\hline
\end{tabular}

\label{tb:gapratio}
\end{table*}

\subsection{Baseline}
%Before the baselines, we introduce two algorithms that we used to generated the baselines:
The ranking algorithms in our baselines are as follows:

\textbf{Empirical Weight (EW).} This algorithm applies a weighted sum of the feature values with feature weights where the weights were empirically adjusted by engineering experts.  %Note that it has nothing to do with machine learning and is completely from experience and intuition. 

\textbf{Learning to Rank (L2R).} This ranking algorithm learns feature weights by a point-wise learning-to-rank network whose structure is the same as the actor network shown in Figure \ref{fig:ltr} but without message as input. The network is supervised by the user feedback of whether a click/purchase happens on an item.

The main difference among EW, L2R and MA-RDPG is the way to generate the feature weights. In MA-RDPG, feature weights are produced by the actor networks. Some typical ranking features are listed in Table ~\ref{Feature}.

On top of the algorithms, we compared MA-RDPG with three baselines that separately optimized the ranking strategies in the main search and in-shop search: 1) EW+L2R; 2) L2R+EW; 3) L2R+L2R. The first algorithm indicates the one used for the main search, and the second one for the in-shop search.

\subsection{Result}
\subsubsection{Metric}
We reported the relative improvement between the compared model against the model in which EW is deployed on both scenarios (main and in-shop search), EW+EW. The metric, GMV gap, is defined as $\frac{(GMV(x)-GMV(y))}{GMV(y)}$, the relative GMV growth  of a model ($GMV(x)$) compared to the setting of EW+EW ($GMV(y)$).
To make a fair comparison, all the algorithms run seven days in our A/B test system where 3\% users are selected into the test group. The performance is measured in terms of the GMV gap in both search scenarios in these days. The performance for each single scenario is also provided as an indicator so that we can study the correlation between the two scenarios.

\subsubsection{Result Analysis}
The results are shown in Table ~\ref{tb:gapratio} and we made the following observations:

\textbf{First},
our MA-RDPG performs much better than all the baselines which are equipped with L2R or empirical weights. In particular, MA-RDPG outperforms L2R+L2R which is a strong model currently using by Taobao, but L2R+L2R independently optimizes the ranking strategies in main search and in-shop search. It justifies that the collaboration between scenarios truly improves the overall GMV.

\textbf{Second},
with MA-RDPG, the GMV of in-shop search is improved significantly while the main search agent maintains comparable GMVs.
The reason is that the traffic from the main search to the in-shop search is much more than that from the in-shop search to the main search (25.46\% vs.9.12\%). Thus, the in-shop search agent is benefited more by receiving messages from the main search agent.

\textbf{Third},
the results of L2R+EW further validate our motivations that the two scenarios should cooperate with each other because improving GMV in the main search hurts that in the in-shop search.

\begin{figure}[t]
 \centering
   \includegraphics[width=0.43\textwidth]{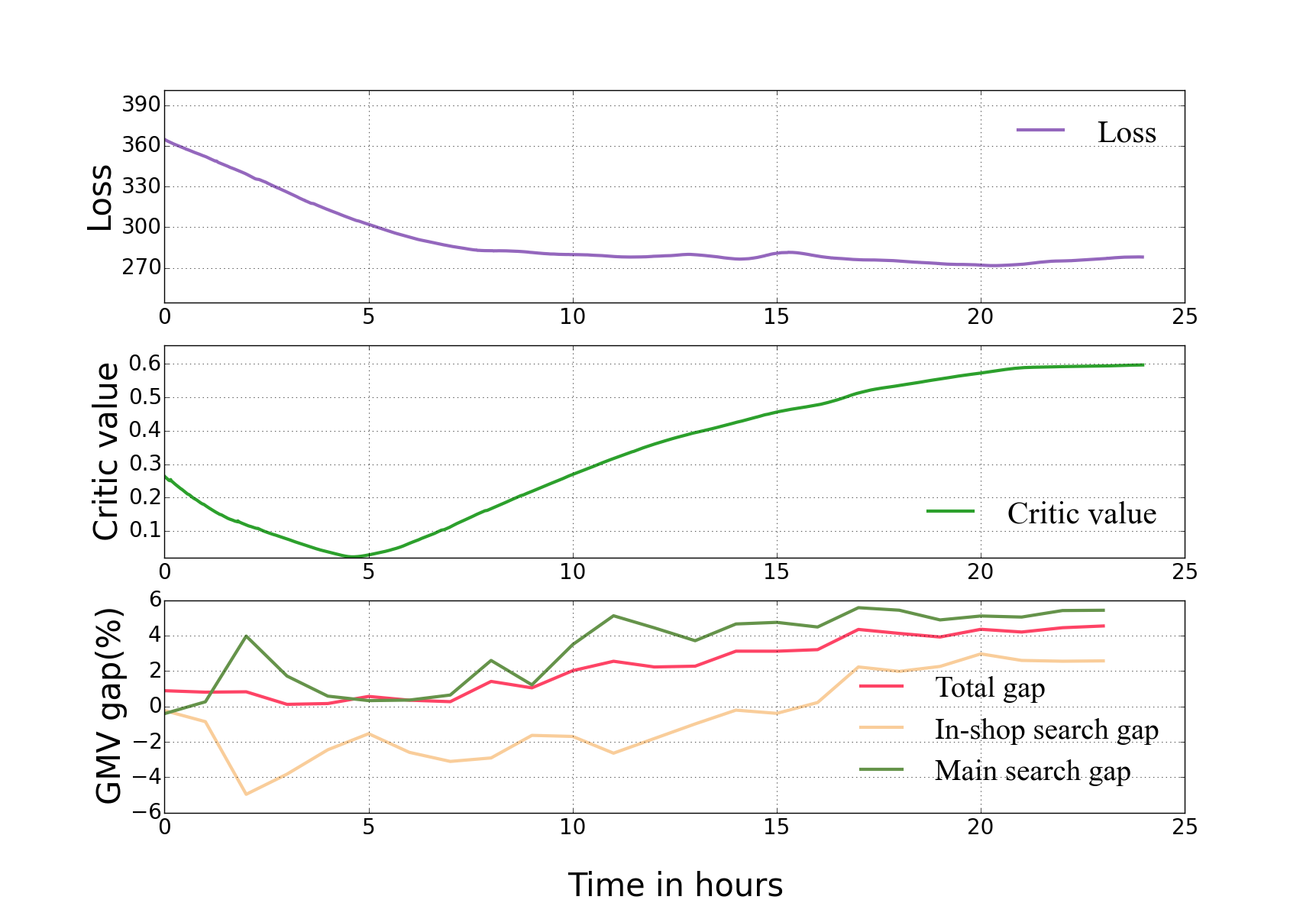}
   %\vspace{-0.1in}
   \caption{Upper/Middle: Learning process of the critic/actor network respectively. Lower: GMV gap against the EW+EW baseline in the online experiments. }
   \label{fig:gmv_gap}
\end{figure}

\begin{figure}[t]
      \includegraphics[width=0.43\textwidth]{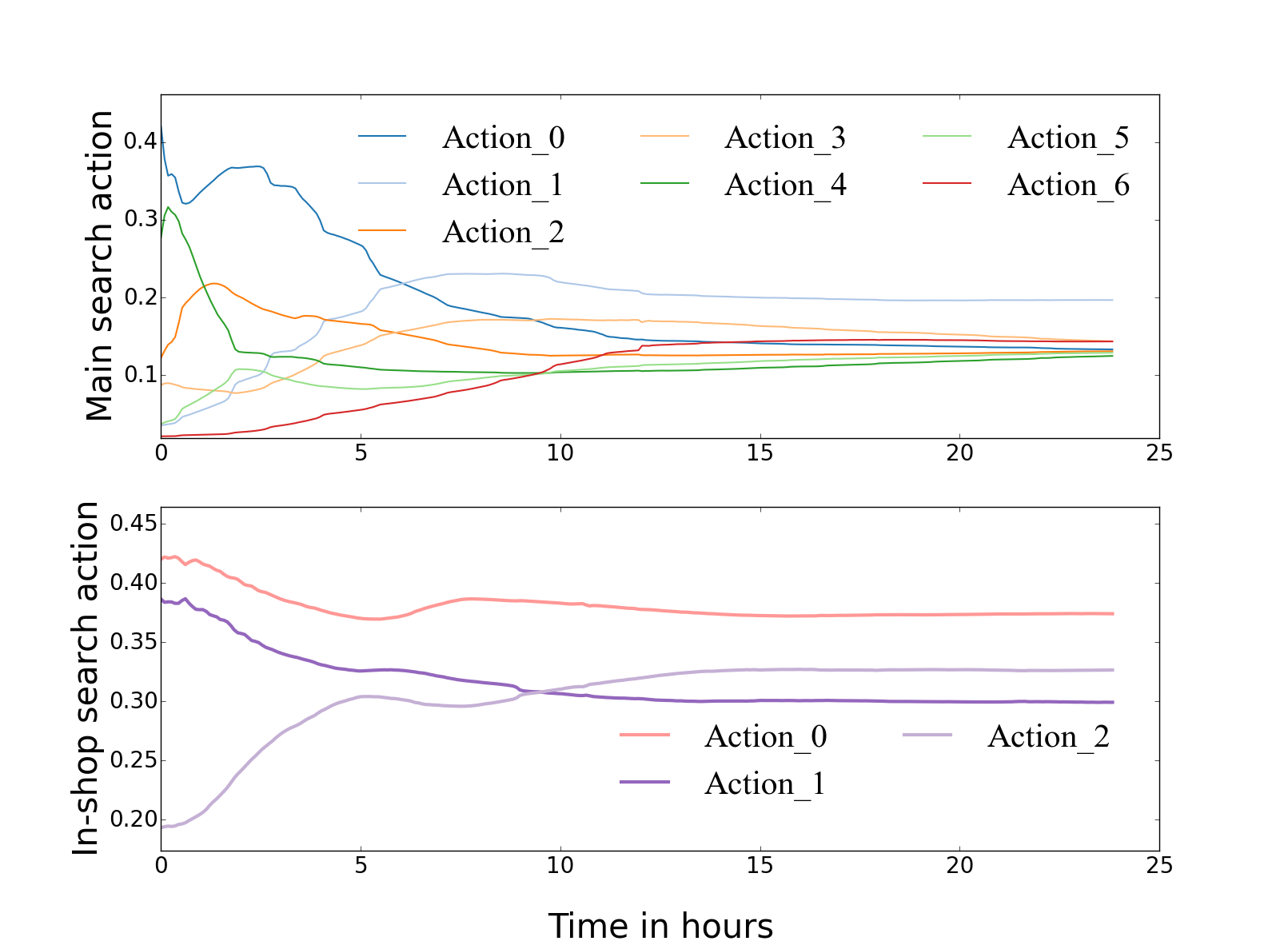}
   %\vspace{-0.1in}
   \caption{The change of main and in-shop search actions. Actions are averaged over the outputs of an actor network within a training batch. }
   \label{fig:time}
\end{figure}

In the lower sub-figure of Figure \ref{fig:gmv_gap} we investigated the stability of MA-RDPG by plotting the mean performance which averages GMV gaps at the same hour within the seven days. It shows that MA-RDPG makes stable and continuous improvement.

\subsubsection{Action Analysis}
As aforementioned, we employed continuous actions for the agents. We thus evaluated how the actions change over time, as shown in Figure \ref{fig:time}. Since each dimension of an action vector is real-valued, we reported the average of the action vectors in a training batch in this experiment.  

The upper sub-figure of Figure \ref{fig:time} depicts how the actions of the main search change over time. \textbf{Action\_1} has the largest value in the main search, corresponding to the feature of Click-Through-Rate (see Table ~\ref{Feature}). This indicates that Click-Through-Rate is the most important feature, in line with the fact that CTR is known as a very important factor. \textbf{Action\_6} is second largest, and it represents the weight of Shop Popularity (but not Item Popularity). However, this feature used to be a weak one in L2R, but plays a much more important role in our experiment than expected. With this feature, the main search can direct more traffic to the in-shop search by providing products from popular shops.

The change of the actions of the in-shop search is illustrated in the lower figure of Figure \ref{fig:time}.  \textbf{Action\_0} is the most influential feature, which represents the weight of Sales Volume (see Table ~\ref{Feature}). More popular items seem to be bought more. 
Though the values of the actions varied dramatically at the early stage, they converged to stable values after about 15 hours' training. This is accordant with the loss and critic value curves as shown in Figure \ref{fig:gmv_gap} which shows that the critic and actor networks converged finally.
\begin{figure}[t]
      \includegraphics[width=0.32\textwidth]{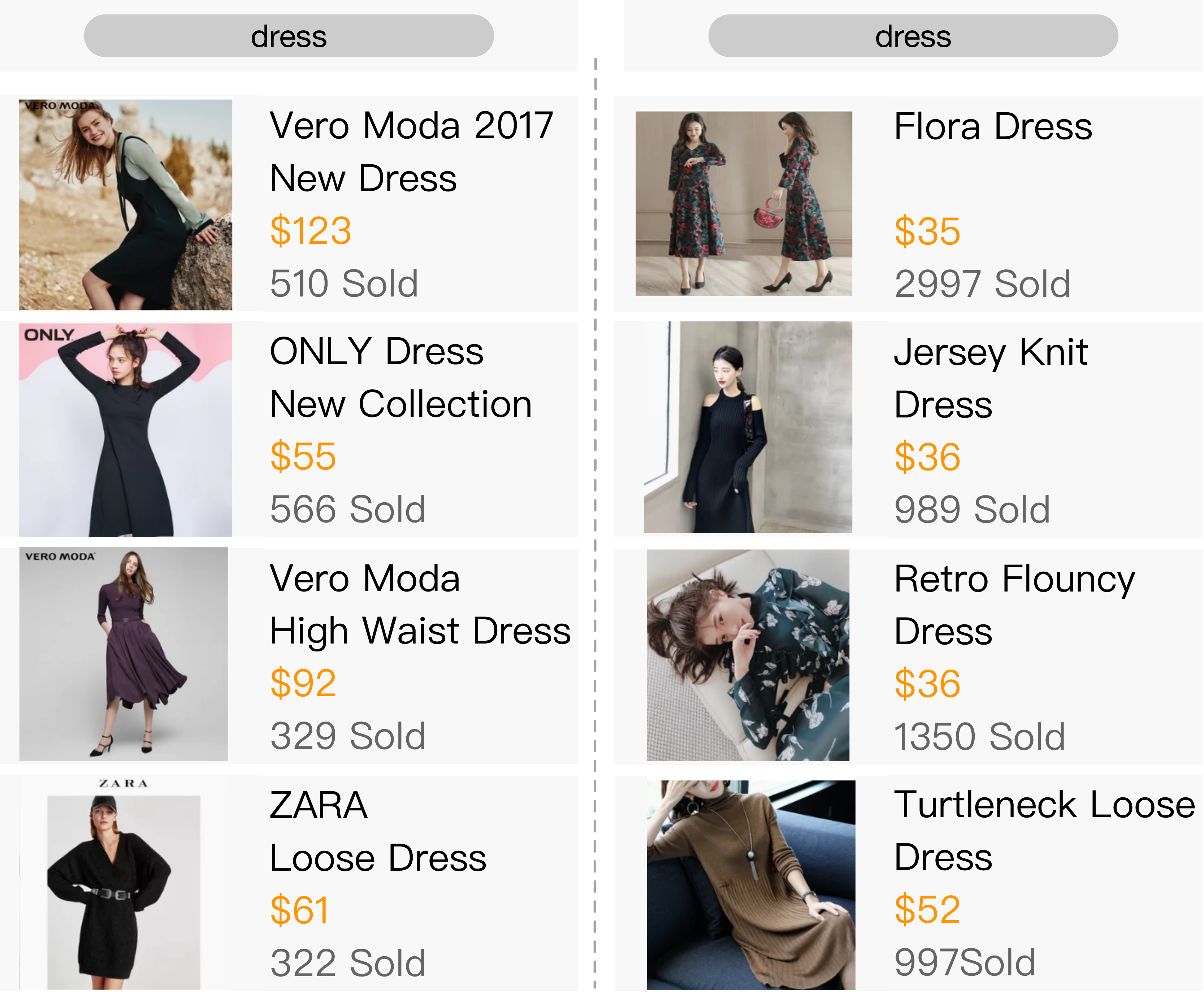} 
   %\vspace{-0.1in}
   \caption{Search result comparison. The left is by MA-RDPG and the right by L2R+L2R.}
   \label{fig:case}
\end{figure}
\subsection{Case Study}
In this subsection we further analyzed a case on how main search and in-shop search cooperate by MA-RDPG. %Due to the high variability of the online environment, we only focused on some typical situations and shown the differences of the ranking items between MA-RDPG and L2R+L2R. 
The case illustrates how the main search helps the in-shop search, thereby targeting more future overall rewards. We simulated a scene from user log like this: a young woman with strong purchase intent clicked some items of skirt which are expensive and have low conversion rates, then she queried ``dress'' in the main search. The results returned by the two models are shown in Figure \ref{fig:case}. Obviously, the results of MA-RDPG are with lower sales (small sold numbers) but with more expensive prices from more branded shops, which makes customers enter the shops with a high probability. By contrast to L2R+L2R, the main search with MA-RDPG ranks items from a global perspective in that it does not only consider its own immediate reward but also the future potential purchase during the in-shop search. 

% 第二个例子是shop search的例子
%For the other case, we selected a situation where a young man wants to buy a refrigerator. At first, he resorted to the main search and received a list of refrigerators. Then, he clicked an item and was provided with a detailed page. The shop of the item is a very large electronics mall so he entered the shop page in which the item list is ranked by the in-shop search engine. As shown in Figure \ref{figmuti}(b), MA-RDPG presented more refrigerators in the front of the page than L2R+L2R due to better understanding of user's intent. Our model provided better in-shop search results for the customer because it takes into account user's context in the main search.
%\begin{figure}[t]
% \centering
%   \includegraphics[width=0.3\textwidth]{IMG_0924_7.png}
%   %\vspace{-0.1in}
%   \caption{Comparison of the main search results provided by %MA-RDPG (left) and L2R\&L2R (right).}
%   \label{fig:case1}
% %\vspace{-0.1in}	
%\end{figure}
%\begin{figure}[t]
% \centering
%   \includegraphics[width=0.3\textwidth]{IMG_0937_2.png}
%   %\vspace{-0.1in}
%   \caption{Comparison of the in-shop search results provided by %MA-RDPG (left) and L2R\&L2R (right).}
%   \label{fig:case2}
% %\vspace{-0.1in}	
%\end{figure}

\section{Related Work}
Ranking is a fundamental problem in many applications such as searching, recommendation, and advertising systems. A good ranking strategy can significantly improve user experience and the performance of the applications.
Learning to rank (L2R) is one typical genre of popular ranking algorithms~\cite{liu2009learning,li2014learning}, and has been widely applied to E-commerce search~\cite{karmaker2017application}, web search~\cite{macdonald2013whens,azzopardi2016advances}, recommendation system~\cite{shi2010list,shi2016learning}. 
The diversity of ranking results is a common issue studied by the community, as addressed by maximal marginal relevance~\cite{carbonell1998use}, topic representation ~\cite{he2012combining}, and reinforcement learning~\cite{DBLP:conf/sigir/XiaXLGZC17}. 
The efficiency issue is another important problem for large online platforms, which can be addressed by feature selection~\cite{geng2007feature,wang2010learning,wang2010ranking} , cascade learning~\cite{viola2001rapid,wang2011cascade,DBLP:conf/kdd/LiuXOS17}, and many other techniques.

Online large platforms generally have multiple ranking scenarios in multiple sub-domains, however,
joint optimization for multi-scenario ranking is rather unexplored.
We cast the problem as a fully cooperative, partially observable multi-agent reinforcement learning problem.
Multi-agent reinforcement learning problem can be grouped into three categories~\cite{busoniu2010multi}: fully cooperative, full competitive, and mixed strategies. Our task is formulated as a fully cooperative problem, which has a long history in multi-agent learning ~\cite{claus1998dynamics,lauer2000algorithm,panait2005cooperative}.
Due to the increased complexity of the environment, recent research efforts are paid to developing deep reinforcement learning (DRL) models. ~\cite{gupta2017cooperative} combined three training schemes with DRL models to make agents collaborate with each other. Counterfactual Multi-Agent  Policy Gradients~\cite{foerster2017counterfactual} uses a centralised critic to estimate a global Q-function. In addition, it uses a counterfactual baseline that marginalises out the action of a single agent, to address the challenges of multi-agent credit assignment. ~\cite{sunehag2017value} introduced a novel additive value-decomposition approach over individual agents instead of learning a shared total reward. However, agents in these three models cannot communicate with each other. Thus, the communication protocols~\cite{foerster2016learning} were proposed to make agents collaborate more easily. The base model of ~\cite{foerster2016learning} is deep Q-Learning, which is not suitable for continuous action space.

Therefore, we propose our own Multi-Agent Recurrent Deterministic Policy Gradient (MA-RDPG) model. %MA-RDPG is formulated within a multi-agent actor-critic framework and is suitable for both discrete and continuous actions.

\section{Conclusion}
In this paper, we present a multi-agent reinforcement learning model, MA-RDPG which employs continuous actions, deterministic policies, and recurrent message encodings. The model can optimize ranking strategies collaboratively for multi-scenario ranking problems. The model consists of a centralized critic, private actors (agents), and a communication component. Actors (agents) work collaboratively in two manners: sharing the same action-value function (the critic) that estimates the future overall rewards, and sending communication messages that encode all historical contexts. The model demonstrates advantages over baselines through online evaluation on an E-commerce platform. 

The proposed model is a general framework which may be applicable to other joint ranking/optimization problems. We leave this as future work.

\section{Acknowledgement}
This work was partly supported by the National Science Foundation of China under grant No.61272227/61332007.
\newpage
\bibliographystyle{ACM-Reference-Format}
\bibliography{sample-bibliography} 


\begin{thebibliography}{50}

%%% the Web address.

\ifx \showCODEN    \undefined \def \showCODEN     #1{\unskip}     \fi
\ifx \showDOI      \undefined \def \showDOI       #1{#1}\fi
\ifx \showISBNx    \undefined \def \showISBNx     #1{\unskip}     \fi
\ifx \showISBNxiii \undefined \def \showISBNxiii  #1{\unskip}     \fi
\ifx \showISSN     \undefined \def \showISSN      #1{\unskip}     \fi
\ifx \showLCCN     \undefined \def \showLCCN      #1{\unskip}     \fi
\ifx \shownote     \undefined \def \shownote      #1{#1}          \fi
\ifx \showarticletitle \undefined \def \showarticletitle #1{#1}   \fi
\ifx \showURL      \undefined \def \showURL       {\relax}        \fi
% invisible to TeX
\providecommand\bibfield[2]{#2}
\providecommand\bibinfo[2]{#2}
\providecommand\natexlab[1]{#1}
\providecommand\showeprint[2][]{arXiv:#2}

\bibitem[\protect\citeauthoryear{Azzopardi and Zuccon}{Azzopardi and
  Zuccon}{2016}]%
        {azzopardi2016advances}
\bibfield{author}{\bibinfo{person}{Leif Azzopardi} {and} \bibinfo{person}{Guido
  Zuccon}.} \bibinfo{year}{2016}\natexlab{}.
\newblock \showarticletitle{Advances in formal models of search and search
  behaviour}. In \bibinfo{booktitle}{\emph{Proceedings of the 2016 ACM on
  International Conference on the Theory of Information Retrieval}}. ACM,
  \bibinfo{pages}{1--4}.
\newblock


\bibitem[\protect\citeauthoryear{Burges, Shaked, Renshaw, Lazier, Deeds,
  Hamilton, and Hullender}{Burges et~al\mbox{.}}{2005}]%
        {burges2005learning}
\bibfield{author}{\bibinfo{person}{Chris Burges}, \bibinfo{person}{Tal Shaked},
  \bibinfo{person}{Erin Renshaw}, \bibinfo{person}{Ari Lazier},
  \bibinfo{person}{Matt Deeds}, \bibinfo{person}{Nicole Hamilton}, {and}
  \bibinfo{person}{Greg Hullender}.} \bibinfo{year}{2005}\natexlab{}.
\newblock \showarticletitle{Learning to rank using gradient descent}. In
  \bibinfo{booktitle}{\emph{Proceedings of the 22nd international conference on
  Machine learning}}. ACM, \bibinfo{pages}{89--96}.
\newblock


\bibitem[\protect\citeauthoryear{Burges, Ragno, and Le}{Burges
  et~al\mbox{.}}{2007}]%
        {burges2007learning}
\bibfield{author}{\bibinfo{person}{Christopher~J Burges},
  \bibinfo{person}{Robert Ragno}, {and} \bibinfo{person}{Quoc~V Le}.}
  \bibinfo{year}{2007}\natexlab{}.
\newblock \showarticletitle{Learning to rank with nonsmooth cost functions}. In
  \bibinfo{booktitle}{\emph{NIPS}}. \bibinfo{pages}{193--200}.
\newblock


\bibitem[\protect\citeauthoryear{Busoniu, Babuska, and De~Schutter}{Busoniu
  et~al\mbox{.}}{2008}]%
        {busoniu2008comprehensive}
\bibfield{author}{\bibinfo{person}{Lucian Busoniu}, \bibinfo{person}{Robert
  Babuska}, {and} \bibinfo{person}{Bart De~Schutter}.}
  \bibinfo{year}{2008}\natexlab{}.
\newblock \showarticletitle{A comprehensive survey of multiagent reinforcement
  learning}.
\newblock \bibinfo{journal}{\emph{IEEE Transactions on Systems, Man, And
  Cybernetics-Part C: Applications and Reviews, 38 (2), 2008}}
  (\bibinfo{year}{2008}).
\newblock


\bibitem[\protect\citeauthoryear{Busoniu, Babu{\v{s}}ka, and
  De~Schutter}{Busoniu et~al\mbox{.}}{2010}]%
        {busoniu2010multi}
\bibfield{author}{\bibinfo{person}{Lucian Busoniu}, \bibinfo{person}{Robert
  Babu{\v{s}}ka}, {and} \bibinfo{person}{Bart De~Schutter}.}
  \bibinfo{year}{2010}\natexlab{}.
\newblock \showarticletitle{Multi-agent reinforcement learning: An overview}.
\newblock \bibinfo{journal}{\emph{Innovations in multi-agent systems and
  applications-1}}  \bibinfo{volume}{310} (\bibinfo{year}{2010}),
  \bibinfo{pages}{183--221}.
\newblock


\bibitem[\protect\citeauthoryear{Cao, Qin, Liu, Tsai, and Li}{Cao
  et~al\mbox{.}}{2007}]%
        {cao2007learning}
\bibfield{author}{\bibinfo{person}{Zhe Cao}, \bibinfo{person}{Tao Qin},
  \bibinfo{person}{Tie-Yan Liu}, \bibinfo{person}{Ming-Feng Tsai}, {and}
  \bibinfo{person}{Hang Li}.} \bibinfo{year}{2007}\natexlab{}.
\newblock \showarticletitle{Learning to rank: from pairwise approach to
  listwise approach}. In \bibinfo{booktitle}{\emph{ICML}}. ACM,
  \bibinfo{pages}{129--136}.
\newblock


\bibitem[\protect\citeauthoryear{Carbonell and Goldstein}{Carbonell and
  Goldstein}{1998}]%
        {carbonell1998use}
\bibfield{author}{\bibinfo{person}{Jaime Carbonell} {and} \bibinfo{person}{Jade
  Goldstein}.} \bibinfo{year}{1998}\natexlab{}.
\newblock \showarticletitle{The use of MMR, diversity-based reranking for
  reordering documents and producing summaries}. In
  \bibinfo{booktitle}{\emph{the 21st ACM SIGIR}}. ACM,
  \bibinfo{pages}{335--336}.
\newblock


\bibitem[\protect\citeauthoryear{Clark}{Clark}{2015}]%
        {clark2015}
\bibfield{author}{\bibinfo{person}{Jack Clark}.} \bibinfo{year}{Retrieved 28
  October2015}\natexlab{}.
\newblock \showarticletitle{Google Turning Its Lucrative Web Search Over to AI
  Machines}.
\newblock \bibinfo{journal}{\emph{Bloomberg Business}}
  (\bibinfo{year}{Retrieved 28 October2015}).
\newblock


\bibitem[\protect\citeauthoryear{Claus and Boutilier}{Claus and
  Boutilier}{1998}]%
        {claus1998dynamics}
\bibfield{author}{\bibinfo{person}{Caroline Claus} {and} \bibinfo{person}{Craig
  Boutilier}.} \bibinfo{year}{1998}\natexlab{}.
\newblock \showarticletitle{The dynamics of reinforcement learning in
  cooperative multiagent systems}.
\newblock \bibinfo{journal}{\emph{AAAI/IAAI}}  \bibinfo{volume}{1998}
  (\bibinfo{year}{1998}), \bibinfo{pages}{746--752}.
\newblock


\bibitem[\protect\citeauthoryear{Covington, Adams, and Sargin}{Covington
  et~al\mbox{.}}{2016}]%
        {Covington2016Deep}
\bibfield{author}{\bibinfo{person}{Paul Covington}, \bibinfo{person}{Jay
  Adams}, {and} \bibinfo{person}{Emre Sargin}.}
  \bibinfo{year}{2016}\natexlab{}.
\newblock \showarticletitle{Deep Neural Networks for YouTube Recommendations}.
  In \bibinfo{booktitle}{\emph{ACM Conference on Recommender Systems}}.
  \bibinfo{pages}{191--198}.
\newblock


\bibitem[\protect\citeauthoryear{Davidson and Deneckere}{Davidson and
  Deneckere}{1986}]%
        {davidson1986long}
\bibfield{author}{\bibinfo{person}{Carl Davidson} {and}
  \bibinfo{person}{Raymond Deneckere}.} \bibinfo{year}{1986}\natexlab{}.
\newblock \showarticletitle{Long-run competition in capacity, short-run
  competition in price, and the Cournot model}.
\newblock \bibinfo{journal}{\emph{The Rand Journal of Economics}}
  (\bibinfo{year}{1986}), \bibinfo{pages}{404--415}.
\newblock


\bibitem[\protect\citeauthoryear{Foerster, Assael, de~Freitas, and
  Whiteson}{Foerster et~al\mbox{.}}{2016}]%
        {foerster2016learning}
\bibfield{author}{\bibinfo{person}{Jakob Foerster}, \bibinfo{person}{Yannis~M
  Assael}, \bibinfo{person}{Nando de Freitas}, {and} \bibinfo{person}{Shimon
  Whiteson}.} \bibinfo{year}{2016}\natexlab{}.
\newblock \showarticletitle{Learning to communicate with deep multi-agent
  reinforcement learning}. In \bibinfo{booktitle}{\emph{Advances in Neural
  Information Processing Systems}}. \bibinfo{pages}{2137--2145}.
\newblock


\bibitem[\protect\citeauthoryear{Foerster, Farquhar, Afouras, Nardelli, and
  Whiteson}{Foerster et~al\mbox{.}}{2017}]%
        {foerster2017counterfactual}
\bibfield{author}{\bibinfo{person}{Jakob Foerster}, \bibinfo{person}{Gregory
  Farquhar}, \bibinfo{person}{Triantafyllos Afouras}, \bibinfo{person}{Nantas
  Nardelli}, {and} \bibinfo{person}{Shimon Whiteson}.}
  \bibinfo{year}{2017}\natexlab{}.
\newblock \showarticletitle{Counterfactual Multi-Agent Policy Gradients}.
\newblock \bibinfo{journal}{\emph{arXiv preprint arXiv:1705.08926}}
  (\bibinfo{year}{2017}).
\newblock


\bibitem[\protect\citeauthoryear{Geng, Liu, Qin, and Li}{Geng
  et~al\mbox{.}}{2007}]%
        {geng2007feature}
\bibfield{author}{\bibinfo{person}{Xiubo Geng}, \bibinfo{person}{Tie-Yan Liu},
  \bibinfo{person}{Tao Qin}, {and} \bibinfo{person}{Hang Li}.}
  \bibinfo{year}{2007}\natexlab{}.
\newblock \showarticletitle{Feature selection for ranking}. In
  \bibinfo{booktitle}{\emph{Proceedings of the 30th annual international ACM
  SIGIR conference on Research and development in information retrieval}}. ACM,
  \bibinfo{pages}{407--414}.
\newblock


\bibitem[\protect\citeauthoryear{Gey}{Gey}{1994}]%
        {gey1994inferring}
\bibfield{author}{\bibinfo{person}{Fredric~C Gey}.}
  \bibinfo{year}{1994}\natexlab{}.
\newblock \showarticletitle{Inferring probability of relevance using the method
  of logistic regression}. In \bibinfo{booktitle}{\emph{Proceedings of the 17th
  annual international ACM SIGIR conference on Research and development in
  information retrieval}}. Springer-Verlag New York, Inc.,
  \bibinfo{pages}{222--231}.
\newblock


\bibitem[\protect\citeauthoryear{Gupta, Egorov, and Kochenderfer}{Gupta
  et~al\mbox{.}}{2017}]%
        {gupta2017cooperative}
\bibfield{author}{\bibinfo{person}{Jayesh~K Gupta}, \bibinfo{person}{Maxim
  Egorov}, {and} \bibinfo{person}{Mykel Kochenderfer}.}
  \bibinfo{year}{2017}\natexlab{}.
\newblock \showarticletitle{Cooperative multiagent control using deep
  reinforcement learning}. In \bibinfo{booktitle}{\emph{Proceedings of the
  Adaptive and Learning Agents workshop (at AAMAS 2017)}}.
\newblock


\bibitem[\protect\citeauthoryear{Gupta, Pathak, and Mitra}{Gupta
  et~al\mbox{.}}{2015}]%
        {Gupta2015Complementary}
\bibfield{author}{\bibinfo{person}{Saurabh Gupta}, \bibinfo{person}{Sayan
  Pathak}, {and} \bibinfo{person}{Bivas Mitra}.}
  \bibinfo{year}{2015}\natexlab{}.
\newblock \bibinfo{booktitle}{\emph{Complementary Usage of Tips and Reviews for
  Location Recommendation in Yelp}}.
\newblock \bibinfo{publisher}{Springer International Publishing}. 1003--1003
  pages.
\newblock


\bibitem[\protect\citeauthoryear{Hausknecht and Stone}{Hausknecht and
  Stone}{2015}]%
        {hausknecht2015deep}
\bibfield{author}{\bibinfo{person}{Matthew Hausknecht} {and}
  \bibinfo{person}{Peter Stone}.} \bibinfo{year}{2015}\natexlab{}.
\newblock \showarticletitle{Deep recurrent q-learning for partially observable
  mdps}.
\newblock  (\bibinfo{year}{2015}).
\newblock


\bibitem[\protect\citeauthoryear{He, Hollink, and de~Vries}{He
  et~al\mbox{.}}{2012}]%
        {he2012combining}
\bibfield{author}{\bibinfo{person}{Jiyin He}, \bibinfo{person}{Vera Hollink},
  {and} \bibinfo{person}{Arjen de Vries}.} \bibinfo{year}{2012}\natexlab{}.
\newblock \showarticletitle{Combining implicit and explicit topic
  representations for result diversification}. In \bibinfo{booktitle}{\emph{the
  35th ACM SIGIR}}. ACM, \bibinfo{pages}{851--860}.
\newblock


\bibitem[\protect\citeauthoryear{Heess, Wayne, Silver, Lillicrap, Erez, and
  Tassa}{Heess et~al\mbox{.}}{2015}]%
        {heess2015learning}
\bibfield{author}{\bibinfo{person}{Nicolas Heess}, \bibinfo{person}{Gregory
  Wayne}, \bibinfo{person}{David Silver}, \bibinfo{person}{Tim Lillicrap},
  \bibinfo{person}{Tom Erez}, {and} \bibinfo{person}{Yuval Tassa}.}
  \bibinfo{year}{2015}\natexlab{}.
\newblock \showarticletitle{Learning continuous control policies by stochastic
  value gradients}. In \bibinfo{booktitle}{\emph{NIPS}}.
  \bibinfo{pages}{2944--2952}.
\newblock


\bibitem[\protect\citeauthoryear{Hochreiter and Schmidhuber}{Hochreiter and
  Schmidhuber}{1997}]%
        {hochreiter1997long}
\bibfield{author}{\bibinfo{person}{Sepp Hochreiter} {and}
  \bibinfo{person}{J{\"u}rgen Schmidhuber}.} \bibinfo{year}{1997}\natexlab{}.
\newblock \showarticletitle{Long short-term memory}.
\newblock \bibinfo{journal}{\emph{Neural computation}} \bibinfo{volume}{9},
  \bibinfo{number}{8} (\bibinfo{year}{1997}), \bibinfo{pages}{1735--1780}.
\newblock


\bibitem[\protect\citeauthoryear{Hu, Wellman, et~al\mbox{.}}{Hu
  et~al\mbox{.}}{1998}]%
        {hu1998multiagent}
\bibfield{author}{\bibinfo{person}{Junling Hu}, \bibinfo{person}{Michael~P
  Wellman}, {et~al\mbox{.}}} \bibinfo{year}{1998}\natexlab{}.
\newblock \showarticletitle{Multiagent reinforcement learning: theoretical
  framework and an algorithm.}. In \bibinfo{booktitle}{\emph{ICML}},
  Vol.~\bibinfo{volume}{98}. \bibinfo{pages}{242--250}.
\newblock


\bibitem[\protect\citeauthoryear{Karmaker~Santu, Sondhi, and
  Zhai}{Karmaker~Santu et~al\mbox{.}}{2017}]%
        {karmaker2017application}
\bibfield{author}{\bibinfo{person}{Shubhra~Kanti Karmaker~Santu},
  \bibinfo{person}{Parikshit Sondhi}, {and} \bibinfo{person}{ChengXiang Zhai}.}
  \bibinfo{year}{2017}\natexlab{}.
\newblock \showarticletitle{On Application of Learning to Rank for E-Commerce
  Search}. In \bibinfo{booktitle}{\emph{Proceedings of the 40th International
  ACM SIGIR Conference on Research and Development in Information Retrieval}}.
  ACM, \bibinfo{pages}{475--484}.
\newblock


\bibitem[\protect\citeauthoryear{Kenthapadi, Kenthapadi, and
  Kenthapadi}{Kenthapadi et~al\mbox{.}}{2017}]%
        {Kenthapadi2017LiJAR}
\bibfield{author}{\bibinfo{person}{Krishnaram Kenthapadi},
  \bibinfo{person}{Krishnaram Kenthapadi}, {and} \bibinfo{person}{Krishnaram
  Kenthapadi}.} \bibinfo{year}{2017}\natexlab{}.
\newblock \showarticletitle{LiJAR: A System for Job Application Redistribution
  towards Efficient Career Marketplace}. In \bibinfo{booktitle}{\emph{ACM
  SIGKDD International Conference on Knowledge Discovery and Data Mining}}.
  \bibinfo{pages}{1397--1406}.
\newblock


\bibitem[\protect\citeauthoryear{Konda and Tsitsiklis}{Konda and
  Tsitsiklis}{2000}]%
        {konda2000actor}
\bibfield{author}{\bibinfo{person}{Vijay~R Konda} {and} \bibinfo{person}{John~N
  Tsitsiklis}.} \bibinfo{year}{2000}\natexlab{}.
\newblock \showarticletitle{Actor-critic algorithms}. In
  \bibinfo{booktitle}{\emph{Advances in neural information processing
  systems}}. \bibinfo{pages}{1008--1014}.
\newblock


\bibitem[\protect\citeauthoryear{Lauer and Riedmiller}{Lauer and
  Riedmiller}{2000}]%
        {lauer2000algorithm}
\bibfield{author}{\bibinfo{person}{Martin Lauer} {and} \bibinfo{person}{Martin
  Riedmiller}.} \bibinfo{year}{2000}\natexlab{}.
\newblock \showarticletitle{An algorithm for distributed reinforcement learning
  in cooperative multi-agent systems}. In \bibinfo{booktitle}{\emph{In
  Proceedings of the Seventeenth International Conference on Machine
  Learning}}. Citeseer.
\newblock


\bibitem[\protect\citeauthoryear{Li}{Li}{2014}]%
        {li2014learning}
\bibfield{author}{\bibinfo{person}{Hang Li}.} \bibinfo{year}{2014}\natexlab{}.
\newblock \showarticletitle{Learning to rank for information retrieval and
  natural language processing}.
\newblock \bibinfo{journal}{\emph{Synthesis Lectures on Human Language
  Technologies}} \bibinfo{volume}{7}, \bibinfo{number}{3}
  (\bibinfo{year}{2014}), \bibinfo{pages}{1--121}.
\newblock


\bibitem[\protect\citeauthoryear{Li, Chu, Langford, and Wang}{Li
  et~al\mbox{.}}{2011}]%
        {li2011unbiased}
\bibfield{author}{\bibinfo{person}{Lihong Li}, \bibinfo{person}{Wei Chu},
  \bibinfo{person}{John Langford}, {and} \bibinfo{person}{Xuanhui Wang}.}
  \bibinfo{year}{2011}\natexlab{}.
\newblock \showarticletitle{Unbiased offline evaluation of
  contextual-bandit-based news article recommendation algorithms}. In
  \bibinfo{booktitle}{\emph{Proceedings of the fourth ACM international
  conference on Web search and data mining}}. ACM, \bibinfo{pages}{297--306}.
\newblock


\bibitem[\protect\citeauthoryear{Li, Wu, and Burges}{Li et~al\mbox{.}}{2008}]%
        {li2008mcrank}
\bibfield{author}{\bibinfo{person}{Ping Li}, \bibinfo{person}{Qiang Wu}, {and}
  \bibinfo{person}{Christopher~J Burges}.} \bibinfo{year}{2008}\natexlab{}.
\newblock \showarticletitle{Mcrank: Learning to rank using multiple
  classification and gradient boosting}. In \bibinfo{booktitle}{\emph{Advances
  in neural information processing systems}}. \bibinfo{pages}{897--904}.
\newblock


\bibitem[\protect\citeauthoryear{Lillicrap, Hunt, Pritzel, Heess, Erez, Tassa,
  Silver, and Wierstra}{Lillicrap et~al\mbox{.}}{2015}]%
        {lillicrap2015continuous}
\bibfield{author}{\bibinfo{person}{Timothy~P Lillicrap},
  \bibinfo{person}{Jonathan~J Hunt}, \bibinfo{person}{Alexander Pritzel},
  \bibinfo{person}{Nicolas Heess}, \bibinfo{person}{Tom Erez},
  \bibinfo{person}{Yuval Tassa}, \bibinfo{person}{David Silver}, {and}
  \bibinfo{person}{Daan Wierstra}.} \bibinfo{year}{2015}\natexlab{}.
\newblock \showarticletitle{Continuous control with deep reinforcement
  learning}.
\newblock \bibinfo{journal}{\emph{arXiv preprint arXiv:1509.02971}}
  (\bibinfo{year}{2015}).
\newblock


\bibitem[\protect\citeauthoryear{Littman}{Littman}{1994}]%
        {littman1994markov}
\bibfield{author}{\bibinfo{person}{Michael~L Littman}.}
  \bibinfo{year}{1994}\natexlab{}.
\newblock \showarticletitle{Markov games as a framework for multi-agent
  reinforcement learning}. In \bibinfo{booktitle}{\emph{Proceedings of the
  eleventh international conference on machine learning}},
  Vol.~\bibinfo{volume}{157}. \bibinfo{pages}{157--163}.
\newblock


\bibitem[\protect\citeauthoryear{Liu, Xiao, Ou, and Si}{Liu
  et~al\mbox{.}}{2017}]%
        {DBLP:conf/kdd/LiuXOS17}
\bibfield{author}{\bibinfo{person}{Shichen Liu}, \bibinfo{person}{Fei Xiao},
  \bibinfo{person}{Wenwu Ou}, {and} \bibinfo{person}{Luo Si}.}
  \bibinfo{year}{2017}\natexlab{}.
\newblock \showarticletitle{Cascade Ranking for Operational E-commerce Search}.
  In \bibinfo{booktitle}{\emph{Proceedings of the 23rd {ACM} {SIGKDD}
  International Conference on Knowledge Discovery and Data Mining, Halifax, NS,
  Canada, August 13 - 17, 2017}}. \bibinfo{pages}{1557--1565}.
\newblock


\bibitem[\protect\citeauthoryear{Liu et~al\mbox{.}}{Liu et~al\mbox{.}}{2009}]%
        {liu2009learning}
\bibfield{author}{\bibinfo{person}{Tie-Yan Liu} {et~al\mbox{.}}}
  \bibinfo{year}{2009}\natexlab{}.
\newblock \showarticletitle{Learning to rank for information retrieval}.
\newblock \bibinfo{journal}{\emph{Foundations and Trends{\textregistered} in
  Information Retrieval}} \bibinfo{volume}{3}, \bibinfo{number}{3}
  (\bibinfo{year}{2009}), \bibinfo{pages}{225--331}.
\newblock


\bibitem[\protect\citeauthoryear{Macdonald, Santos, and Ounis}{Macdonald
  et~al\mbox{.}}{2013}]%
        {macdonald2013whens}
\bibfield{author}{\bibinfo{person}{Craig Macdonald},
  \bibinfo{person}{Rodrygo~LT Santos}, {and} \bibinfo{person}{Iadh Ounis}.}
  \bibinfo{year}{2013}\natexlab{}.
\newblock \showarticletitle{The whens and hows of learning to rank for web
  search}.
\newblock \bibinfo{journal}{\emph{Information Retrieval}} \bibinfo{volume}{16},
  \bibinfo{number}{5} (\bibinfo{year}{2013}), \bibinfo{pages}{584--628}.
\newblock


\bibitem[\protect\citeauthoryear{Mnih, Kavukcuoglu, Silver, Rusu, Veness,
  Bellemare, Graves, Riedmiller, Fidjeland, Ostrovski, et~al\mbox{.}}{Mnih
  et~al\mbox{.}}{2015}]%
        {mnih2015human}
\bibfield{author}{\bibinfo{person}{Volodymyr Mnih}, \bibinfo{person}{Koray
  Kavukcuoglu}, \bibinfo{person}{David Silver}, \bibinfo{person}{Andrei~A
  Rusu}, \bibinfo{person}{Joel Veness}, \bibinfo{person}{Marc~G Bellemare},
  \bibinfo{person}{Alex Graves}, \bibinfo{person}{Martin Riedmiller},
  \bibinfo{person}{Andreas~K Fidjeland}, \bibinfo{person}{Georg Ostrovski},
  {et~al\mbox{.}}} \bibinfo{year}{2015}\natexlab{}.
\newblock \showarticletitle{Human-level control through deep reinforcement
  learning}.
\newblock \bibinfo{journal}{\emph{Nature}} \bibinfo{volume}{518},
  \bibinfo{number}{7540} (\bibinfo{year}{2015}), \bibinfo{pages}{529--533}.
\newblock


\bibitem[\protect\citeauthoryear{Panait and Luke}{Panait and Luke}{2005}]%
        {panait2005cooperative}
\bibfield{author}{\bibinfo{person}{Liviu Panait} {and} \bibinfo{person}{Sean
  Luke}.} \bibinfo{year}{2005}\natexlab{}.
\newblock \showarticletitle{Cooperative multi-agent learning: The state of the
  art}.
\newblock \bibinfo{journal}{\emph{Autonomous agents and multi-agent systems}}
  \bibinfo{volume}{11}, \bibinfo{number}{3} (\bibinfo{year}{2005}),
  \bibinfo{pages}{387--434}.
\newblock


\bibitem[\protect\citeauthoryear{Qin, Zhang, Wang, Liu, Lai, and Li}{Qin
  et~al\mbox{.}}{2007}]%
        {qin2007ranking}
\bibfield{author}{\bibinfo{person}{Tao Qin}, \bibinfo{person}{Xu-Dong Zhang},
  \bibinfo{person}{De-Sheng Wang}, \bibinfo{person}{Tie-Yan Liu},
  \bibinfo{person}{Wei Lai}, {and} \bibinfo{person}{Hang Li}.}
  \bibinfo{year}{2007}\natexlab{}.
\newblock \showarticletitle{Ranking with multiple hyperplanes}. In
  \bibinfo{booktitle}{\emph{Proceedings of the 30th annual international ACM
  SIGIR conference on Research and development in information retrieval}}. ACM,
  \bibinfo{pages}{279--286}.
\newblock


\bibitem[\protect\citeauthoryear{Santamar{\'\i}a, Sutton, and
  Ram}{Santamar{\'\i}a et~al\mbox{.}}{1997}]%
        {santamaria1997experiments}
\bibfield{author}{\bibinfo{person}{Juan~C Santamar{\'\i}a},
  \bibinfo{person}{Richard~S Sutton}, {and} \bibinfo{person}{Ashwin Ram}.}
  \bibinfo{year}{1997}\natexlab{}.
\newblock \showarticletitle{Experiments with reinforcement learning in problems
  with continuous state and action spaces}.
\newblock \bibinfo{journal}{\emph{Adaptive behavior}} \bibinfo{volume}{6},
  \bibinfo{number}{2} (\bibinfo{year}{1997}), \bibinfo{pages}{163--217}.
\newblock


\bibitem[\protect\citeauthoryear{Shi, Ifrim, and Hurley}{Shi
  et~al\mbox{.}}{2016}]%
        {shi2016learning}
\bibfield{author}{\bibinfo{person}{Bichen Shi}, \bibinfo{person}{Georgiana
  Ifrim}, {and} \bibinfo{person}{Neil Hurley}.}
  \bibinfo{year}{2016}\natexlab{}.
\newblock \showarticletitle{Learning-to-rank for real-time high-precision
  hashtag recommendation for streaming news}. In
  \bibinfo{booktitle}{\emph{Proceedings of the 25th International Conference on
  World Wide Web}}. International World Wide Web Conferences Steering
  Committee, \bibinfo{pages}{1191--1202}.
\newblock


\bibitem[\protect\citeauthoryear{Shi, Larson, and Hanjalic}{Shi
  et~al\mbox{.}}{2010}]%
        {shi2010list}
\bibfield{author}{\bibinfo{person}{Yue Shi}, \bibinfo{person}{Martha Larson},
  {and} \bibinfo{person}{Alan Hanjalic}.} \bibinfo{year}{2010}\natexlab{}.
\newblock \showarticletitle{List-wise learning to rank with matrix
  factorization for collaborative filtering}. In
  \bibinfo{booktitle}{\emph{Proceedings of the fourth ACM conference on
  Recommender systems}}. ACM, \bibinfo{pages}{269--272}.
\newblock


\bibitem[\protect\citeauthoryear{Sunehag, Lever, Gruslys, Czarnecki, Zambaldi,
  Jaderberg, Lanctot, Sonnerat, Leibo, Tuyls, et~al\mbox{.}}{Sunehag
  et~al\mbox{.}}{2017}]%
        {sunehag2017value}
\bibfield{author}{\bibinfo{person}{Peter Sunehag}, \bibinfo{person}{Guy Lever},
  \bibinfo{person}{Audrunas Gruslys}, \bibinfo{person}{Wojciech~Marian
  Czarnecki}, \bibinfo{person}{Vinicius Zambaldi}, \bibinfo{person}{Max
  Jaderberg}, \bibinfo{person}{Marc Lanctot}, \bibinfo{person}{Nicolas
  Sonnerat}, \bibinfo{person}{Joel~Z Leibo}, \bibinfo{person}{Karl Tuyls},
  {et~al\mbox{.}}} \bibinfo{year}{2017}\natexlab{}.
\newblock \showarticletitle{Value-Decomposition Networks For Cooperative
  Multi-Agent Learning}.
\newblock \bibinfo{journal}{\emph{arXiv preprint arXiv:1706.05296}}
  (\bibinfo{year}{2017}).
\newblock


\bibitem[\protect\citeauthoryear{Sutton and Barto}{Sutton and Barto}{1998}]%
        {sutton1998reinforcement}
\bibfield{author}{\bibinfo{person}{Richard~S Sutton} {and}
  \bibinfo{person}{Andrew~G Barto}.} \bibinfo{year}{1998}\natexlab{}.
\newblock \bibinfo{booktitle}{\emph{Reinforcement learning: An introduction}}.
  Vol.~\bibinfo{volume}{1}.
\newblock \bibinfo{publisher}{MIT press Cambridge}.
\newblock


\bibitem[\protect\citeauthoryear{Sutton, McAllester, Singh, and Mansour}{Sutton
  et~al\mbox{.}}{2000}]%
        {sutton2000policy}
\bibfield{author}{\bibinfo{person}{Richard~S Sutton}, \bibinfo{person}{David~A
  McAllester}, \bibinfo{person}{Satinder~P Singh}, {and}
  \bibinfo{person}{Yishay Mansour}.} \bibinfo{year}{2000}\natexlab{}.
\newblock \showarticletitle{Policy gradient methods for reinforcement learning
  with function approximation}. In \bibinfo{booktitle}{\emph{NIPS}}.
  \bibinfo{pages}{1057--1063}.
\newblock


\bibitem[\protect\citeauthoryear{Viola and Jones}{Viola and Jones}{2001}]%
        {viola2001rapid}
\bibfield{author}{\bibinfo{person}{Paul Viola} {and} \bibinfo{person}{Michael
  Jones}.} \bibinfo{year}{2001}\natexlab{}.
\newblock \showarticletitle{Rapid object detection using a boosted cascade of
  simple features}. In \bibinfo{booktitle}{\emph{Computer Vision and Pattern
  Recognition, 2001. CVPR 2001. Proceedings of the 2001 IEEE Computer Society
  Conference on}}, Vol.~\bibinfo{volume}{1}. IEEE, \bibinfo{pages}{I--I}.
\newblock


\bibitem[\protect\citeauthoryear{Wang, Lin, and Metzler}{Wang
  et~al\mbox{.}}{2010a}]%
        {wang2010learning}
\bibfield{author}{\bibinfo{person}{Lidan Wang}, \bibinfo{person}{Jimmy Lin},
  {and} \bibinfo{person}{Donald Metzler}.} \bibinfo{year}{2010}\natexlab{a}.
\newblock \showarticletitle{Learning to efficiently rank}. In
  \bibinfo{booktitle}{\emph{Proceedings of the 33rd international ACM SIGIR
  conference on Research and development in information retrieval}}. ACM,
  \bibinfo{pages}{138--145}.
\newblock


\bibitem[\protect\citeauthoryear{Wang, Lin, and Metzler}{Wang
  et~al\mbox{.}}{2011}]%
        {wang2011cascade}
\bibfield{author}{\bibinfo{person}{Lidan Wang}, \bibinfo{person}{Jimmy Lin},
  {and} \bibinfo{person}{Donald Metzler}.} \bibinfo{year}{2011}\natexlab{}.
\newblock \showarticletitle{A cascade ranking model for efficient ranked
  retrieval}. In \bibinfo{booktitle}{\emph{Proceedings of the 34th
  international ACM SIGIR conference on Research and development in Information
  Retrieval}}. ACM, \bibinfo{pages}{105--114}.
\newblock


\bibitem[\protect\citeauthoryear{Wang, Metzler, and Lin}{Wang
  et~al\mbox{.}}{2010b}]%
        {wang2010ranking}
\bibfield{author}{\bibinfo{person}{Lidan Wang}, \bibinfo{person}{Donald
  Metzler}, {and} \bibinfo{person}{Jimmy Lin}.}
  \bibinfo{year}{2010}\natexlab{b}.
\newblock \showarticletitle{Ranking under temporal constraints}. In
  \bibinfo{booktitle}{\emph{Proceedings of the 19th ACM international
  conference on Information and knowledge management}}. ACM,
  \bibinfo{pages}{79--88}.
\newblock


\bibitem[\protect\citeauthoryear{Watkins and Dayan}{Watkins and Dayan}{1992}]%
        {watkins1992q}
\bibfield{author}{\bibinfo{person}{Christopher~JCH Watkins} {and}
  \bibinfo{person}{Peter Dayan}.} \bibinfo{year}{1992}\natexlab{}.
\newblock \showarticletitle{Q-learning}.
\newblock \bibinfo{journal}{\emph{Machine learning}} \bibinfo{volume}{8},
  \bibinfo{number}{3-4} (\bibinfo{year}{1992}), \bibinfo{pages}{279--292}.
\newblock


\bibitem[\protect\citeauthoryear{Xia, Xu, Lan, Guo, Zeng, and Cheng}{Xia
  et~al\mbox{.}}{2017}]%
        {DBLP:conf/sigir/XiaXLGZC17}
\bibfield{author}{\bibinfo{person}{Long Xia}, \bibinfo{person}{Jun Xu},
  \bibinfo{person}{Yanyan Lan}, \bibinfo{person}{Jiafeng Guo},
  \bibinfo{person}{Wei Zeng}, {and} \bibinfo{person}{Xueqi Cheng}.}
  \bibinfo{year}{2017}\natexlab{}.
\newblock \showarticletitle{Adapting Markov Decision Process for Search Result
  Diversification}. In \bibinfo{booktitle}{\emph{Proceedings of the 40th
  International {ACM} {SIGIR} Conference on Research and Development in
  Information Retrieval, Shinjuku, Tokyo, Japan, August 7-11, 2017}}.
  \bibinfo{pages}{535--544}.
\newblock


\bibitem[\protect\citeauthoryear{Yin, Hu, Tang, Daly, Zhou, Ouyang, Chen, Kang,
  Deng, Nobata, et~al\mbox{.}}{Yin et~al\mbox{.}}{2016}]%
        {yin2016ranking}
\bibfield{author}{\bibinfo{person}{Dawei Yin}, \bibinfo{person}{Yuening Hu},
  \bibinfo{person}{Jiliang Tang}, \bibinfo{person}{Tim Daly},
  \bibinfo{person}{Mianwei Zhou}, \bibinfo{person}{Hua Ouyang},
  \bibinfo{person}{Jianhui Chen}, \bibinfo{person}{Changsung Kang},
  \bibinfo{person}{Hongbo Deng}, \bibinfo{person}{Chikashi Nobata},
  {et~al\mbox{.}}} \bibinfo{year}{2016}\natexlab{}.
\newblock \showarticletitle{Ranking relevance in yahoo search}. In
  \bibinfo{booktitle}{\emph{Proceedings of the 22nd ACM SIGKDD International
  Conference on Knowledge Discovery and Data Mining}}. ACM,
  \bibinfo{pages}{323--332}.
\newblock


\end{thebibliography}
\end{document}